\documentclass[11pt]{article}

\usepackage[preprint]{acl}

\usepackage{times}
\usepackage{latexsym}

\usepackage[T1]{fontenc}

\usepackage[utf8]{inputenc}

\usepackage{microtype}

\usepackage{inconsolata}

\usepackage{graphicx}

\usepackage[acronym]{glossaries}
\glsdisablehyper
\newacronym{llm}{LLM}{large language model}
\newacronym{vqlc}{VQLC}{Vector Quantized Latent Concept}
\newacronym{ema}{EMA}{exponentially moving average}
\newacronym{lacoat}{LACOAT}{latent concept attribution}
\newacronym{oom}{OOM}{out-of-memory}
\newacronym{sae}{SAE}{Sparse Autoencoder}
\newacronym{ig}{IG}{integrated gradients}
\newacronym{vqvae}{VQ-VAE}{vector quantized-variational autoencoder}
\newacronym{mlp}{MLP}{multi-layer perceptron}
\newacronym{hpc}{HPC}{high-performance computing}

\usepackage{amsfonts}
\newcommand{\modelM}{\mathbb{M}}
\newcommand{\data}{\mathbb{D}}

\usepackage{caption, subcaption}
\usepackage{enumitem}
\setlist[itemize]{leftmargin=14pt}

\usepackage{booktabs} 
\usepackage{dsfont} 
\usepackage{multirow}

\newenvironment{promptbox}
{\begin{center}\begin{minipage}{0.95\linewidth}\hrule\vspace{4pt}\footnotesize\ttfamily\raggedright}
{\vspace{4pt}\hrule\end{minipage}\end{center}}

\title{Vector Quantized Latent Concepts: A Scalable Alternative to \\Clustering-Based Concept Discovery
\\\hphantom{...}
\\\footnotesize{\textcolor{red}{WARNING: The appendix contains some examples, which may be disturbing to the reader.}}}

\author{
  \textbf{Xuemin Yu\textsuperscript{1}} \quad
  \textbf{Ankur Garg\textsuperscript{2}} \quad
  \textbf{Samira Ebrahimi Kahou\textsuperscript{2}} \quad
  \textbf{Hassan Sajjad\textsuperscript{1}} \\
  \textsuperscript{1}Dalhousie University, Canada \\
  \textsuperscript{2}University of Calgary, Canada
}

\begin{document}
\maketitle

\begin{abstract}
\Glspl{llm} encode rich semantic information in their hidden states, yet it remains difficult to understand what information these internal representations capture.
Latent concepts extracted from hidden states offer a promising 
direction 
for interpreting \glspl{llm}, but existing clustering-based methods face 
a trade-off: hierarchical clustering produces coherent concepts but is limited to small datasets due to its quadratic memory cost,
while K-Means scales efficiently but may yield less semantically coherent concepts. We propose \gls{vqlc}, a discrete concept learning framework that learns a codebook of latent concepts on frozen hidden states. Across 12 dataset-model settings, \gls{vqlc} stays close to K-Means in computational cost, scales better than hierarchical clustering, and remains competitive in faithfulness, 
with the clearest gains on decoder-only models. \Glspl{llm}-based evaluation, qualitative analysis, and a \gls{sae} comparison 
demonstrate that the learned concepts are 
interpretable and task-relevant.


\end{abstract}

\section{Introduction}

Current \glspl{llm} achieve impressive capabilities, but their internal representations remain difficult to interpret, and the relationship between these representations and the model output is often opaque~\citep{Rudin2019Stop, huang2023augmenting, shi2024detectingpretrainingdatalarge, dodge-etal-2021-documenting, sheng-etal-2021-societal}. Prior work often explains individual predictions through input attribution methods, such as \gls{ig}, SmoothGrad, and SHAP~\citep{ribeiro2016should, Sundararajan2017, smilkov2017smoothgrad, lundberg2017unified}.  These methods identify input tokens that are salient to a model's prediction, providing a token-level explanation of prediction saliency. However, token-level explanations provide limited insight into the high-level semantic information encoded in internal representations.


\begin{figure}[t]
    \centering
    \includegraphics[width=0.9\linewidth]{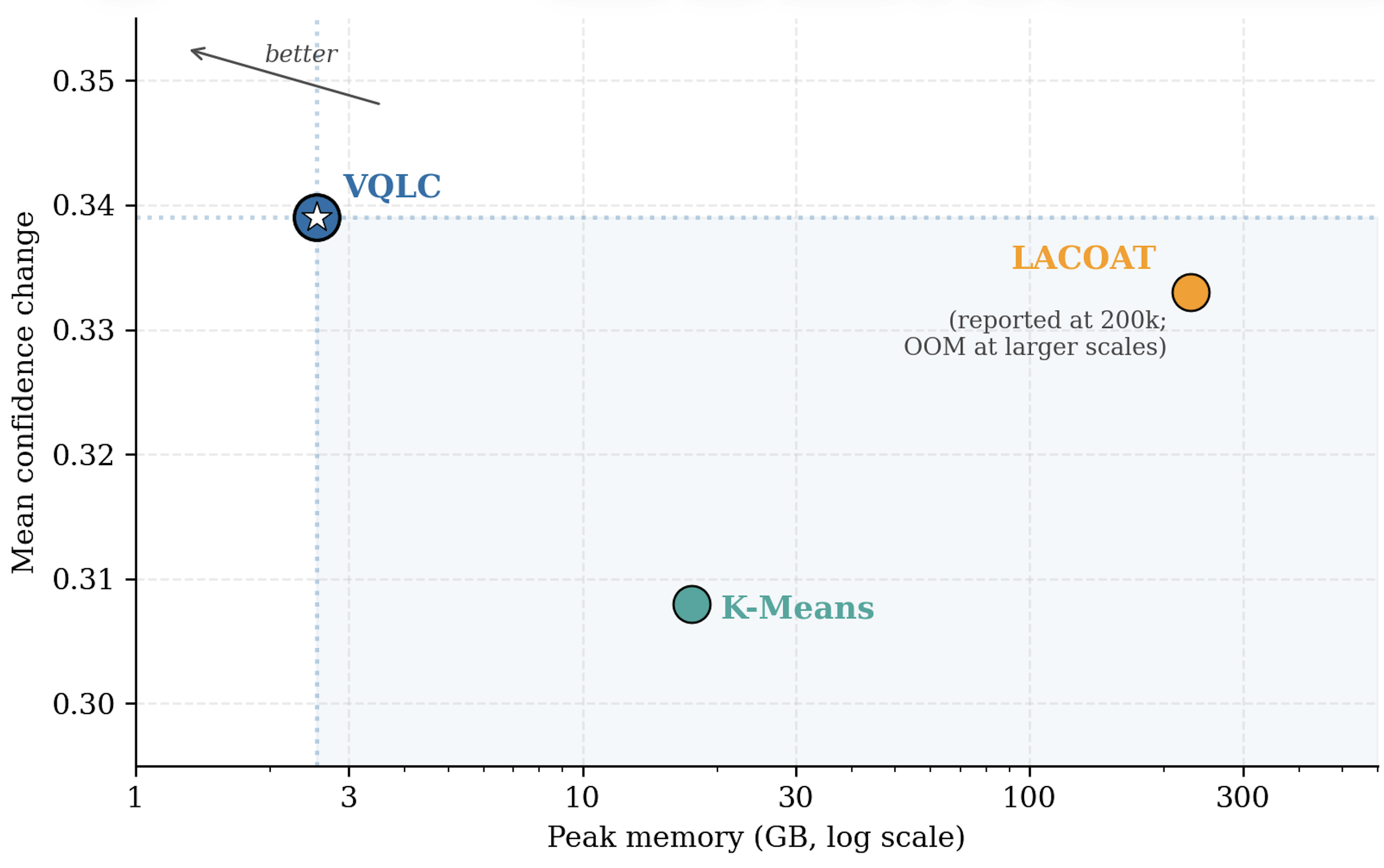}
    \caption{Faithfulness and scalability trade-off: Each point compares a concept-discovery method using the average confidence change across 12 dataset-model settings and the peak memory measured during the scalability evaluation. \Gls{vqlc} achieves the most favorable trade-off, obtaining the highest mean confidence change while requiring the lowest peak memory.\protect\footnotemark}
    \label{fig:pareto}
\end{figure}
\footnotetext{\Gls{vqlc} and K-Means are measured at 500k tokens. \Gls{lacoat} is shown at its largest completed scale at 200k tokens.}

\begin{figure*}[t]
    \centering
    \includegraphics[width=0.85\textwidth]{figures/overview.pdf}
    \caption{Architecture overview: Contextual token representations extracted from a \gls{llm} are encoded and discretized through vector quantization. During training, the decoder reconstructs the token representations. After training, latent concepts are constructed by aggregating encoder outputs with the same code assignments. At inference time, the learned concepts are used to explain the corresponding representations.}
    \label{fig:overview}
\end{figure*}

Another line of work aims to interpret 
hidden representations through latent concepts extracted from contextual representations~\citep{kim2018interpretability, ghorbani_towards_2019, dalvi2022discovering, jourdan2023cockatiel, zhao2024explaining, yu-etal-2024-latent, lam-2024, sharma2025analyzing}. 
The main idea is that a word can have different contextualized representations depending on the context, where each representation captures a different meaning. Representations with similar semantic or functional behavior can be viewed as \textit{concepts}~\citep{dalvi2022discovering}. By organizing representations into latent concepts, concept discovery provides a more structured semantic view of representation space than isolated token-level saliency explanations. Most existing methods discover concepts through post-hoc clustering. For example, \gls{lacoat}~\citep{yu-etal-2024-latent} applies agglomerative hierarchical clustering to token representations. While hierarchical clustering can discover meaningful concepts, it scales poorly as the number of tokens grows. Alternatively, K-Means is more computationally efficient, but it often provides a weaker trade-off between scalability and semantic concept quality.


Vector quantization provides a natural alternative by mapping continuous hidden states to a finite set of learned discrete codes~\citep{Oord2017}. Each representation is assigned to its nearest codebook vector in a shared learned codebook. The reconstruction objective encourages these codebook vectors to preserve information from the original representations while grouping tokens that can be represented by similar codebook vectors. This aligns with the goal of concept discovery, where the codebook provides a finite set of codes that can be interpreted as concepts learned from hidden states. The codebook also keeps the cost of assigning a token to a code constant with respect to dataset size, removing the quadratic dependence that limits hierarchical clustering at scale.

Based on this motivation, we propose \gls{vqlc}, a discrete concept learning framework.
It uses a lightweight residual encoder to map token representations into a code space, a vector quantizer to assign them to their nearest vectors in a learnable codebook, and a residual decoder to reconstruct the original hidden states. After training, the learned codebook maps hidden representations to a finite set of latent concepts. Token representations assigned to the same code are treated as sharing a similar semantic facet.
Our goal is to discover latent concepts that identify the semantic structure encoded in hidden representations.


We evaluate \gls{vqlc} against hierarchical clustering and K-Means across 
models. The results show that \gls{vqlc} remains close to K-Means in computational cost, scales better than hierarchical clustering in a representative large-scale setting, and remains competitive in concept quality, with the clearest gains on decoder-only models. Figure~\ref{fig:pareto} previews this trade-off between faithfulness and scalability. 
Overall, this paper contributes a vector quantized framework for scalable latent concept discovery and a multiple evaluations covering scalability, faithfulness, \glspl{llm}-based judgments, qualitative analysis, and comparisons with \glspl{sae}. 

\section{Problem Formulation}
\label{sec:formulation}
We consider a language model $\modelM$ with layers $\ell \in L$. Given an input instance $s = {w_1, w_2, \dots, w_N}$, let $h_{w_i}^{(\ell)}$ denote the contextual representation of token $w_i$ at layer $\ell$, and $H^{(\ell)}$ denote the set of representations over the training split $\data_{\text{train}}$. Our goal is to learn a discrete codebook $\mathcal{E} = \{e_1, \dots, e_K\}$ such that token representations with similar semantic meaning are mapped to the same discrete codes.
From the resulting assignments, we derive concept vectors $\mathcal{V} = \{v_1, \dots, v_K\}$ by averaging the encoded representations assigned to each code, and corresponding latent concepts  $\mathcal{C} = \{c_1, \dots, c_K\}$, where each $c_k$ consists of a concept vector $v_k$ and its associated tokens. Tokens in the same concept are expected to encode similar semantic facets.


\section{Methodology}
\Gls{vqlc} adapts vector quantization to latent concept discovery on hidden states. As shown in Figure~\ref{fig:overview}, it consists of three main modules: an \textbf{encoder} that maps contextual representations extracted from a chosen layer of frozen \glspl{llm} into a codebook space; a \textbf{vector quantizer} that assigns each encoded representation to its nearest vector in a learnable codebook; and a \textbf{decoder} that reconstructs the original representations from the quantized vectors. During training, the codebook learns to support discrete assignment and reconstruction. After training, we freeze the model and run a concept-construction pass over the training split to derive token-to-code assignments and construct concept vectors. At test time, token representations of a test instance 
are assigned to the learned codebook, and the corresponding latent concepts describe the semantic information encoded in those representations. When the representation chosen for explanation is the one driving model prediction (e.g., the last layer classification token), the assigned concept corresponds to the semantics underlying that prediction.



\subsection{Encoder}
The encoder is a residual \gls{mlp} that maps contextual representations into the code space. It combines a direct linear projection with a lightweight nonlinear correction branch. The linear path preserves a direct dependence on the original hidden states, while the non-linear branch  provides limited reshaping before vector quantization. This design keeps encoded representations close to the original model space while making it 
suitable for stable discrete assignment.


Let $ h_{w_i}^{(\ell)}$ denote the contextual representation of token $w_i$ extracted from layer $\ell$. The encoder produces an output $z_e(w_i)$ defined as:
\begin{subequations}
\label{eq:encoder}
\begin{align}
    u_{w_i}^{(\ell)} &= \operatorname{LN}_{\mathrm{in}}\!\left(h_{w_i}^{(\ell)}\right), \\
    r_{w_i}^{(\ell)} &= W_2\,\operatorname{GELU}\!\left(W_1 u_{w_i}^{(\ell)} + b_1\right) + b_2, \\
    z_e(w_i) &= \operatorname{LN}_{\mathrm{out}}\!\left(W_{\mathrm{proj}} h_{w_i}^{(\ell)} + b_{\mathrm{proj}} + r_{w_i}^{(\ell)}\right),
\end{align}
\end{subequations}
where $W_{\mathrm{proj}} \in \mathbb{R}^{d_c \times d}$ denotes the linear transformation into the code space, $W_1 \in \mathbb{R}^{m \times d}$ and $W_2 \in \mathbb{R}^{d_c \times m}$ are the parameters of the nonlinear residual branch, and $\operatorname{LN}_{\mathrm{in}}$ and $\operatorname{LN}_{\mathrm{out}}$ denote layer normalization. In our experiments, we set $d_c = d$ and 
a hidden dimension of $m = 128$. 
Appendix~\ref{app:encoder} provides an ablation study of the residual encoder.

\subsection{Vector Quantizer}

The vector quantizer maintains a learnable codebook $\mathcal{E} = \{e_1, \dots, e_K\}$, where each $e_j \in \mathbb{R}^{d_c}$, and assigns each encoder output to its nearest codebook vector. These assignments are then used to organize token representations into discrete groups for latent concept construction.

\paragraph{Codebook Initialization}
\label{sec:codebook_initializaton}

The codebook is initialized using encoder outputs extracted from the training split. We first remove near-duplicate candidates by rounding each coordinate to four decimal places and then deduplicating identical rounded vectors. We select $K$ initial vectors with a greedy farthest-first traversal: the first vector is chosen as the candidate farthest from the candidate mean, and each following vector maximizes its distance to the nearest already selected vector. This initialization spreads the initial codebook vectors across diverse regions of the encoded representation space rather than concentrating them in dense local regions. Additional initialization details are provided in Appendix~\ref{app:methodology}. Appendix~\ref{app:initialization} compares the farthest-first initialization against random initialization.

\paragraph{Code Assignment}
During quantization, we compute the squared Euclidean distance between each encoder output $z_e(w_i)$ and every codebook vector:

\begin{equation}  
    D(z_e(w_i), e_j) = \left\| z_e(w_i) - e_j \right\|_2^2.
\end{equation}
Each encoder output is then assigned to its nearest codebook vector:
\begin{subequations}
\begin{align}
    j^{*} &= \arg\min_j D(z_e(w_i), e_j), \\
    z_q(w_i) &= e_{j^{*}}.
\end{align}
\end{subequations}
During training, we use a straight-through estimator so that the forward pass uses the quantized representation while gradients continue to flow through the encoder. At inference time, token representations are assigned deterministically using the learned codebook.

\paragraph{Codebook Learning}
We update the codebook using the \gls{ema}-based method proposed by~\citet{kaiser2018}, which provides stable and smooth updates compared to gradient-based approaches. For each codebook vector $e_j$, \gls{ema} maintains both an accumulated usage frequency $n_j$ and a moving average of the encoder outputs assigned to it $m_j$.

The usage frequency of each vector $e_j$ is calculated as:
\begin{align}
    \label{eq1}
        n_j \leftarrow \lambda n_j + (1 - \lambda) \sum_{i} \mathds{1}\left[z_q\left(w_i\right)=e_j\right],
\end{align}
where $z_q(w_i)$ denotes the codebook vector assigned to token $w_i$, $\mathds{1}$ is an indicator function, and $\lambda$ is a decay parameter.

The corresponding codebook vector $e_j$ is updated towards the average of its assigned encoder outputs:
\begin{subequations}
\begin{align}
    m_j &\leftarrow \lambda m_j+(1-\lambda) \sum_i \mathds{1}\left[z_q\left(w_i\right)=e_j\right] z_e(w_i), \\
    e_j &\leftarrow \frac{m_j}{n_j},
\end{align}
\end{subequations}
where $z_e(w_i)$ is the encoder output for token $w_i$. The decay parameter $\lambda$ controls how much the update relies on the previous codebook state, and we set $\lambda = 0.99$ in all experiments. 

Although \gls{ema} stabilizes codebook learning, some codes may remain persistently unused during training. We 
maintain an inactivity counter for each code and recover codes that do not receive any assignments for a fixed patience window. When recovery is triggered, we select encoder outputs with the largest assignment errors in the current batch and use them to reinitialize eligible dead codes. 
Appendix~\ref{app:methodology} gives details on the dead code recovery, and reports its ablation study in Appendix~\ref{app:deadcode_recovery}.



\paragraph{Latent Concept Construction}
After training, we use the trained \gls{vqlc} components and run a final concept-construction pass over the training split. The 
EMA codebook is then used to assign each encoded token representation to a discrete code index. For each code $k$, we derive a concept vector $v_k$ by averaging the encoded representations assigned to that code:
\begin{align}
   v_k = \frac{1}{|A_k|}\sum_{z_e(w_i) \in A_k} z_e(w_i),
\end{align}
where $A_k$ denotes the set of encoded token representations assigned to code $k$. We use $v_k$ from this concept construction pass rather than the \gls{ema} codebook vector $e_k$ as a concept vector. The codebook vector $e_k$ is updated during training and reflects an exponentially weighted training history. In contrast, $v_k$ is computed in a final pass with the frozen encoder and directly summarizes the representations assigned to that code for interpretation. 

We then define a latent concept explanation $c_k$ as the combination of the concept vector $v_k$ and the tokens assigned to $e_k$. At test time, token representations from a new input instance are assigned to the learned codebook. We use the latent concept assigned to each input token to explain the information encoded in its representation. 


\subsection{Decoder}
The decoder reconstructs the original contextual representations from the quantized representation. It consists of a direct linear reconstruction path and a nonlinear correction branch. This design preserves a simple reconstruction route from the codebook space while allowing the decoder to model residual nonlinear structure.

Let $z_q(w_i) \in \mathbb{R}^{d_c}$ denote the quantized representation assigned to token $w_i$. The decoder output $\hat{h}_{w_i}^{(\ell)} \in \mathbb{R}^{d}$ is defined as:
\begin{subequations}
    \label{eq:decoder}
    \begin{align}
        u_q(w_i) &= \operatorname{LN} \! \left(z_q(w_i)\right), \\
        r_q(w_i) &= W_2\, \operatorname{GELU}\!\left(W_1 u_q(w_i) + b_1\right) + b_2, \\
        \hat{h}_{w_i}^{(\ell)} &= W_{\mathrm{rec}} z_q(w_i) + b_{\mathrm{rec}} + r_q(w_i),
    \end{align}
\end{subequations}
where $W_{\mathrm{rec}} \in \mathbb{R}^{d \times d_c}$ denotes the direct linear reconstruction from the code space to the original hidden-state space, and $W_1 \in \mathbb{R}^{m \times d_c}$ and $W_2 \in \mathbb{R}^{d \times m}$ are the parameters of the nonlinear correction branch. The layer normalization is applied before the nonlinear branch to make the decoder less sensitive to variations in the scale of the quantized representations.

The direct linear reconstruction path captures the dominant structure needed to recover the original hidden states, while the nonlinear branch complements this reconstruction path by modeling residual nonlinear structure in the reconstruction.

\subsection{Training Objective}
The overall training objective consists of two components, as shown in Equation~\ref{eq:objective}. The \textbf{reconstruction loss} trains the decoder to reconstruct the original contextual representations from the quantized vectors. Because gradients propagate back to both the decoder and encoder, this objective encourages the encoder to produce representations that preserve contextual information while remaining compatible with vector quantization. The \textbf{commitment loss} encourages the encoder outputs to stay close to their assigned codebook vectors, which stabilizes discrete assignment during training. The optimizer updates only the encoder and decoder parameters. The codebook is updated through EMA rather than direct gradient steps.

\begin{align}
    \label{eq:objective}
    \mathcal{L} &= \mathcal{L}_{\textbf{rec}} + \beta \mathcal{L}_{\textbf{commit}}\\
    \mathcal{L}_{\textbf{rec}} &= \left\|h_{w_i}^{(\ell)} - \hat{h}_{w_i}^{(\ell)}\right\|_2^2\\
    \mathcal{L}_{\textbf{commit}} &= \left\|z_e(w_i) - \operatorname{sg}\!\left[z_q(w_i)\right] \right\|_2^2,
\end{align}
where $\operatorname{sg}$ is the stop-gradient operator, which prevents gradient updates on the codebook. $\beta$ controls the strength of the commitment constraint. We set $\beta = 0.25$ in our experiments. Appendix~\ref{app:commit} provides a sensitivity analysis for the commitment weight.

\section{Experiment Setup}
\paragraph{Data}
We use three sequence classification tasks: ERASER Movie Reviews~\citep{eraser_sst}, Jigsaw Toxicity~\citep{cjadams_2017}, and AG News~\citep{gulli2005ag}. Appendix~\ref{app:dataset} provides detailed dataset information.

\paragraph{Models}

We evaluate on two fine-tuned encoder models, BERT-base-cased~\citep{devlin2018bert} and RoBERTa~\citep{liu2019roberta}, and two decoder-only \glspl{llm}, Llama-2-7B-chat-hf~\citep{touvron2023llama} and Qwen2.5-3B~\citep{qwen2.5}, in a zero-shot prompting setting. For comparison with \gls{sae}, we use pretrained \glspl{sae} matched to the corresponding model: Qwen-Scope~\citep{qwen_scope} for the Qwen3.5-2B base model~\citep{qwen3.5}, and Gemma-Scope~\citep{gemma_scope} for Gemma-3-4B-IT~\citep{gemma_2025}. Fine-tuning performance and hardware details are reported in Appendices~\ref{app:models} and~\ref{app:setup}.

\begin{figure}[t]
    \centering
    \includegraphics[width=\linewidth]{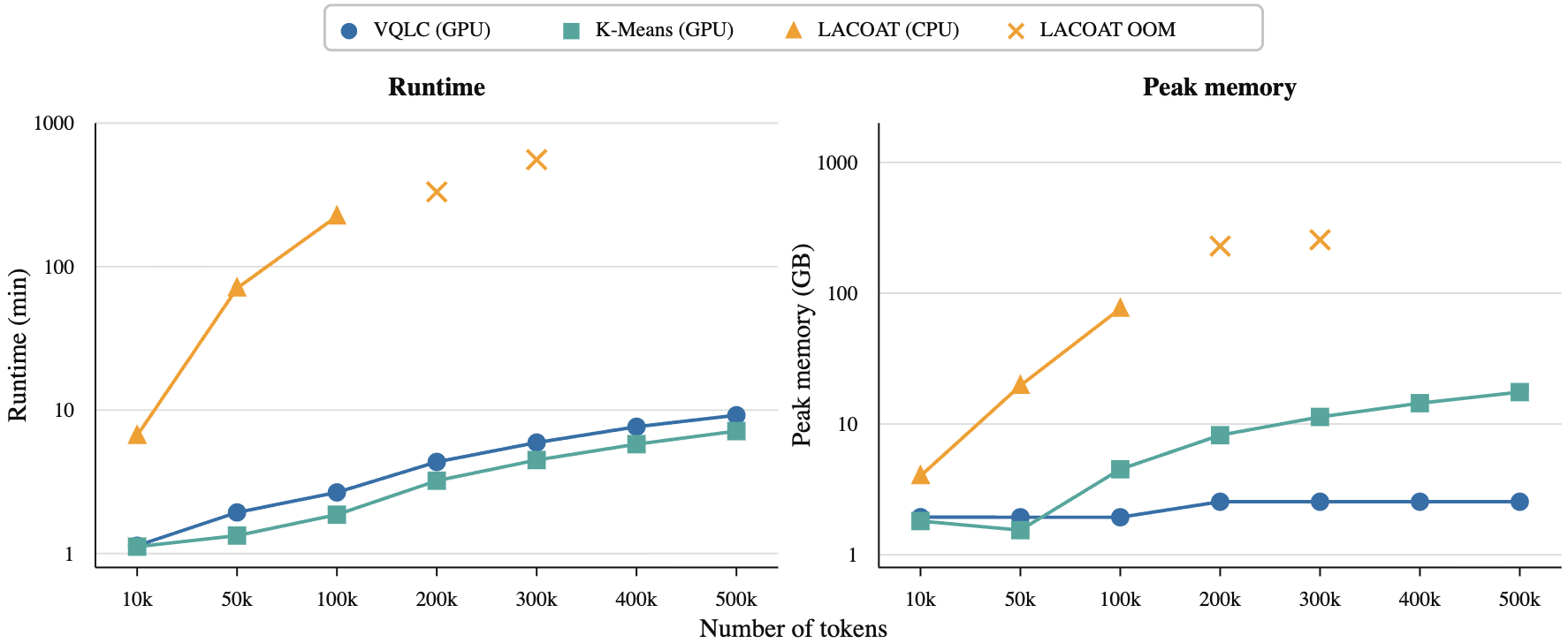}
    \caption{Scalability experiment: runtime and peak memory usage as the number of token representations increases. Dataset: ERASER movie; Model: Qwen.}
    \label{fig:scalability}
\end{figure}

\paragraph{Representation extraction}
We use NeuroX toolkit~\citep{dalvi-etal-2023-neurox} to extract last-layer token activations, following prior work showing that the last layer tends to contain the most task-aligned representations~\citep{ansuini2019intrinsic, roeder2021linear, yu-etal-2024-latent}.
The task decision representation is the 
classification token (e.g. \texttt{[CLS]}) for encoder-based models and the final token representation for decoder-only models. Appendix~\ref{app:dataset} provides data processing details.


\paragraph{Baselines}
To test whether \gls{vqlc} can preserve concept quality while improving scalability, we compare it with two clustering-based baselines: 
\begin{itemize}[leftmargin=*,parsep=1pt,topsep=-2pt, itemsep=2pt]
    \item \textbf{\gls{lacoat}} \citep{yu-etal-2024-latent} uses agglomerative hierarchical clustering to discover latent concepts and trains a classifier to map tokens to concepts. Latent concept vectors are computed by averaging token representations within each cluster. 
    \item \textbf{K-Means} applies K-Means clustering directly to token representations and assigns tokens to the nearest concept vector using cosine similarity at inference time. This baseline tests whether \gls{vqlc} can improve concept quality over scalable centroid-based methods.
\end{itemize}
Because both \gls{vqlc} and K-Means are sensitive to initialization, we run experiments with three seeds $\{0, 42, 999\}$. We also include \glspl{sae} as a complementary comparison to contrast different forms of explanations. Full \gls{vqlc} hyperparameter settings are provided in Appendix~\ref{app:hyperparameters}.


\section{Evaluation}
We evaluate \gls{vqlc} against both K-Means and \gls{lacoat} along four axes: scalability, faithfulness, \glspl{llm}-based evaluation, and qualitative analysis. We additionally compare \gls{vqlc} with \glspl{sae} to highlight how \gls{vqlc} differs from mechanistic feature-based explanation methods. Our main question is whether \gls{vqlc} can serve as a scalable alternative to clustering-based concept discovery without
compromising concept quality.
Figure~\ref{fig:pareto} summarizes this central trade-off. Across 12 dataset-model settings, \gls{vqlc} maintains lower peak memory usage than both clustering baselines as the number of tokens grows, while also achieving the highest mean faithfulness confidence change.
We next examine each evaluation axis in detail.

\begin{table}[t]
 \centering
 \resizebox{\columnwidth}{!}{%
     \begin{tabular}{llccc}
     \toprule
     \textbf{Dataset} & \textbf{Method} & \textbf{RoBERTa} & \textbf{LLaMA} & \textbf{Qwen} \\
     \midrule
     \multirow{3}{*}{\textbf{AG News}}
     & \gls{vqlc}   & \textbf{0.726 $\pm$ 0.003} & \textbf{0.028 $\pm$ 0.000} & 0.195 $\pm$ 0.004 \\
     & K-Means      & 0.718 $\pm$ 0.000          & 0.010 $\pm$ 0.000          & \textbf{0.203 $\pm$ 0.000} \\
     & \gls{lacoat} & 0.722          & 0.026          & 0.196 \\
     \midrule
     \multirow{3}{*}{\textbf{ERASER}}
     & \gls{vqlc}   & 0.493 $\pm$ 0.002          & \textbf{0.060 $\pm$ 0.000} & \textbf{0.090 $\pm$ 0.004} \\
     & K-Means      & \textbf{0.499 $\pm$ 0.000} & 0.023 $\pm$ 0.001          & 0.063 $\pm$ 0.000 \\
     & \gls{lacoat} & 0.484          & 0.041         & 0.063 \\
     \midrule
     \multirow{3}{*}{\textbf{Jigsaw}}
     & \gls{vqlc}   & 0.466 $\pm$ 0.001          & \textbf{0.123 $\pm$ 0.004} & \textbf{0.168 $\pm$ 0.016} \\
     & K-Means      & \textbf{0.483 $\pm$ 0.000} & 0.050 $\pm$ 0.003          & 0.157 $\pm$ 0.000 \\
     & \gls{lacoat} & 0.481          & 0.080          & 0.167 \\
     \bottomrule
     \end{tabular}
 }
 \caption{Faithfulness evaluations across datasets and models: Confidence change after removing the assigned concept direction by orthogonal projection. \textbf{Higher is better ($\uparrow$)}. The best results are \textbf{bolded}. Datasets: AG News, Jigsaw, ERASER movie; Models: RoBERTa, Qwen, LLaMA.} 
 \label{tab:faithfulness}
\end{table}

\subsection{Scalability Evaluation}
\label{sec:scalability}

We evaluate scalability on the ERASER movie dataset using the Qwen model. We scale the number of training token representations 
from 10k to 500k, where
each representation is 2048-dimensional. Due to the difference in execution regimes of baselines and \gls{vqlc}, we report peak GPU memory for \gls{vqlc} and K-Means, and peak CPU resident memory for \gls{lacoat}.


Figure~\ref{fig:scalability} shows that \gls{vqlc} and K-Means both scale substantially better than \gls{lacoat}. \Gls{vqlc} increases from about 1 minute at $10$k tokens to about 9 minutes at $500$k tokens, while K-Means increases from about 1 minute to about 7 minutes over the same range. \Gls{lacoat} already requires about 3.5 hours at 100k tokens. The memory results show the clearest difference: \gls{vqlc} remains nearly constant, increasing from $1.94$GB at $10$k tokens to about $2.54$GB at $500$k tokens. K-Means grows from $1.81$GB to $17.52$GB. \Gls{lacoat} reaches $228.99$GB at $200$k tokens before failing at larger scales. This behavior is consistent with the methods' computational structure. \Gls{vqlc} uses mini-batches with a fixed-size codebook, while hierarchical clustering requires pairwise token similarity computations, resulting in quadratic growth with the number of tokens. K-Means avoids this quadratic dependence, but its clustering procedure 
becomes increasingly memory-intensive as the number of tokens grows.

\subsection{Faithfulness Evaluation}
\label{sec:faithfulness}

We hypothesize that if a latent concept encodes task-relevant information 
in the task-decision representation, then removing its direction from that representation should lead to a larger change in model output. We test this by 
ablating the concept direction through orthogonal projection and comparing the perturbed output with the original output. Appendix~\ref{app:faithfulness} provides details on the projection procedure. We report confidence change, which measures the change in prediction confidence after projection, and additionally report predicted label changes in Appendix~\ref{app:faithfulness}.

Table~\ref{tab:faithfulness} reports the confidence change for RoBERTa, Qwen, and LLaMA across the three datasets. \Gls{vqlc} yields the largest confidence change in 6 out of 9 cases. \Gls{vqlc} achieves its strongest gains in decoder-only models, outperforming the baselines in 5 out of 6 Qwen and LLaMA settings. These results suggest that \gls{vqlc} is particularly effective at identifying task-relevant concept directions in decoder-only models while remaining competitive with clustering-based baselines in encoder-based settings. The 
prediction change results and the BERT comparison are reported in Appendix~\ref{app:faithfulness} (Table~\ref{tab:faithfulness_bert} and Table~\ref{tab:faithfulness_prediction}).

\begin{table}[t]
 \centering

 \resizebox{\columnwidth}{!}{%
     \begin{tabular}{llccc}
     \toprule
     \multicolumn{5}{l}{\textbf{Top: Average Rank ($\downarrow$)}} \\
     \cmidrule(l){1-5}
     \textbf{Dataset} & \textbf{Method} & \textbf{RoBERTa} & \textbf{LLaMA} & \textbf{Qwen} \\
     \midrule
     \multirow{3}{*}{\textbf{AG News}}
     & \gls{vqlc}   & 2.198 $\pm$ 0.129          & 2.082 $\pm$ 0.016          & \textbf{1.999 $\pm$ 0.063} \\
     & K-Means      & 2.011 $\pm$ 0.078          & \textbf{1.186 $\pm$ 0.053} & 2.046 $\pm$ 0.061 \\
     & \gls{lacoat} & \textbf{1.773 $\pm$ 0.026} & 1.914 $\pm$ 0.011          & 2.074 $\pm$ 0.061 \\
     \midrule
     \multirow{3}{*}{\textbf{ERASER}}
     & \gls{vqlc}   & \textbf{1.821 $\pm$ 0.076} & 2.027 $\pm$ 0.075          & \textbf{1.915 $\pm$ 0.083} \\
     & K-Means      & 2.139 $\pm$ 0.096          & 2.259 $\pm$ 0.140          & 2.135 $\pm$ 0.085 \\
     & \gls{lacoat} & 2.052 $\pm$ 0.092          & \textbf{1.659 $\pm$ 0.073} & 1.992 $\pm$ 0.039 \\
     \midrule
     \multirow{3}{*}{\textbf{Jigsaw}}
     & \gls{vqlc}   & \textbf{1.839 $\pm$ 0.070} & 1.632 $\pm$ 0.093          & \textbf{1.878 $\pm$ 0.106} \\
     & K-Means      & 2.247 $\pm$ 0.067          & 2.796 $\pm$ 0.085          & 2.104 $\pm$ 0.096 \\
     & \gls{lacoat} & 2.010 $\pm$ 0.058          & \textbf{1.576 $\pm$ 0.047} & 2.065 $\pm$ 0.053 \\
     \bottomrule
     \end{tabular}
 }
 
 \resizebox{\columnwidth}{!}{%
     \begin{tabular}{llccc}
     \toprule
     \multicolumn{5}{l}{\textbf{Bottom: Mean Kendall's $W$ ($\uparrow$)}} \\
     \cmidrule(l){1-5}
     \textbf{Dataset} & & \textbf{RoBERTa} & \textbf{LLaMA} & \textbf{Qwen} \\
     \midrule
     \multicolumn{2}{l}{\textbf{AG News}} & 0.744 $\pm$ 0.036 & 0.300 $\pm$ 0.036 & 0.741 $\pm$ 0.022 \\
     \multicolumn{2}{l}{\textbf{ERASER}} & 0.665 $\pm$ 0.025 & 0.617 $\pm$ 0.032 & 0.752 $\pm$ 0.047 \\
     \multicolumn{2}{l}{\textbf{Jigsaw}} & 0.699 $\pm$ 0.042 & 0.715 $\pm$ 0.024 & 0.778 $\pm$ 0.023 \\
     \bottomrule
     \end{tabular}
 }
 
 \caption{\Glspl{llm}-based evaluation: \textbf{Top:} Average rank values for each method across datasets and models. 
 \textbf{Bottom:} Mean Kendall's $W$ scores over the three-method rankings, measuring inter-\glspl{llm} agreement.}
 \label{tab:judge_results}

\end{table}

\subsection{\glspl{llm}-Based Evaluation}
 
Following the growing use of LLM-as-a-judge evaluation in \gls{llm} research~\citep{zheng2023judging,li2024llms,shi-etal-2025-judging}, we use multiple \glspl{llm} to judge how well each method's discovered concepts align with the model's output. For each test instance, we provide the sentence, the ground-truth label, the predicted label, and the concept contents generated by \gls{vqlc}, \gls{lacoat}, and K-Means to the \gls{llm} judges. Prompt templates are provided in Appendix~\ref{app:llm}. The \gls{llm} judges assign ranks from 1 to 3, where lower ranks indicate better alignment. To mitigate potential position bias, we randomly shuffle the order in which the three candidate concept explanations are presented. Ties are allowed.


We use three \glspl{llm}: Claude Haiku, Gemini Flash, and DeepSeek~\citep{liu2024deepseek}, and evaluate 50 samples 
for each of the 12 dataset-model settings, for a total of 600 test instances and 1,800 individual \gls{llm} rankings before agreement filtering. 
For each sample, final ranks are determined by 
majority vote across the three evaluators. Samples without an agreement from at least two judges 
are excluded. We report the average rank over the resolved samples and measure inter-LLM agreement using Kendall's $W$, computed per sample over the three method rankings and 
averaged within each setting. Appendix~\ref{app:llm} gives the aggregation details.



Table~\ref{tab:judge_results} reports the average rank (top) and the agreement scores (bottom). \Gls{vqlc} obtains the lowest average rank in 8 of the 12 dataset-model settings. \Gls{lacoat} is best in 3 settings, and K-Means is best in only 1 setting. These results suggest that \gls{vqlc} generally yields more task-aligned latent concepts. Agreement scores are generally moderate to strong, although agreement weakens in AG News with the LLaMA model setting. The BERT results are reported in Appendix~\ref{app:llm}.

\begin{table}[t]
 \centering
 \setlength{\tabcolsep}{4pt}
 \resizebox{\columnwidth}{!}{%
 \begin{tabular}{llccc}
 \toprule
 \textbf{Dataset} & \textbf{Method} & \textbf{Confidence Change} & \textbf{\# Concepts} & \textbf{Active Rate} \\
 \midrule
 \multirow{2}{*}{\textbf{AG News}}
 & \gls{vqlc} & \textbf{0.444} & 399  & \textbf{0.890} \\
 & \gls{sae}-concept top1  & 0.338 & 1834 & 0.375          \\
 \midrule
 \multirow{2}{*}{\textbf{ERASER}}
 & \gls{vqlc} & \textbf{0.296} & 398  & \textbf{0.739} \\
 & \gls{sae}-concept top1  & 0.058  & 601  & 0.216          \\
 \midrule
 \multirow{2}{*}{\textbf{Jigsaw}}
 & \gls{vqlc} & \textbf{0.304} & 398  & \textbf{0.779} \\
 & \gls{sae}-concept top1  & 0.093 & 918  & 0.199          \\
 \bottomrule
 \end{tabular}
 }
 \caption{Comparison between \gls{vqlc} and top-1 \gls{sae} features on the Qwen model. 
 We report faithfulness results, the size of the learned concept inventory, and the concept activation rate on the test set.}
 \label{tab:vqlc_sae}
\end{table}

\subsection{Qualitative Evaluation}

\begin{figure*}[t]
    \centering

    \begin{subfigure}[b]{0.32\textwidth}
        \centering
        \includegraphics[width=\linewidth]{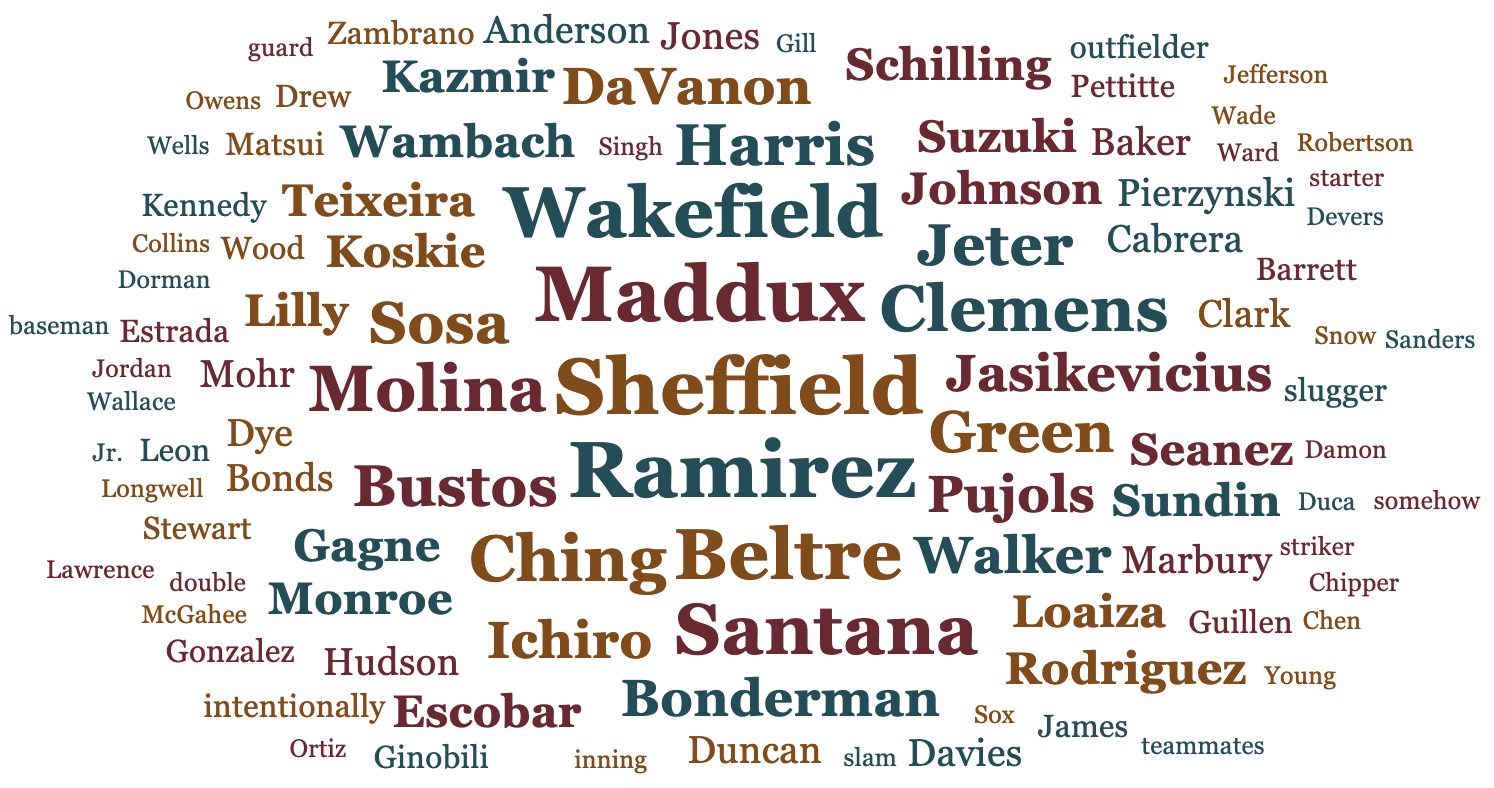} 
        \caption{Baseball Players (Sports)}
        \label{fig:baseball}
    \end{subfigure}
    \hfill 
    \begin{subfigure}[b]{0.32\textwidth}
        \centering
        \includegraphics[width=\linewidth]{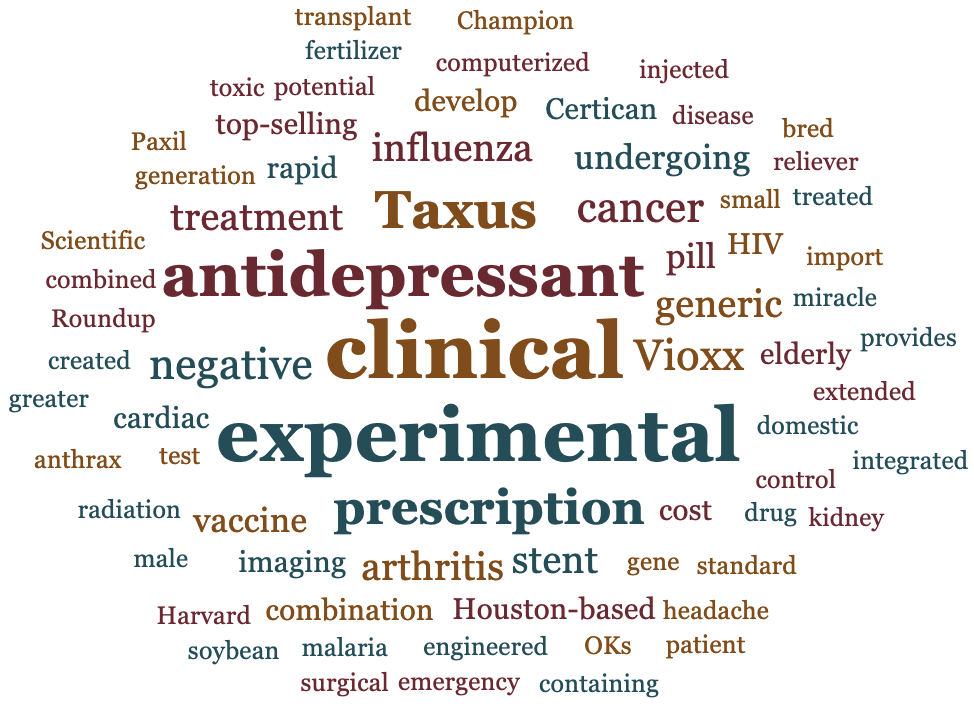} 
        \caption{Clinical Drugs (Sci/Tech)}
        \label{fig:clinical}
    \end{subfigure}
    \hfill
    \begin{subfigure}[b]{0.32\textwidth}
        \centering
        \includegraphics[width=\linewidth]{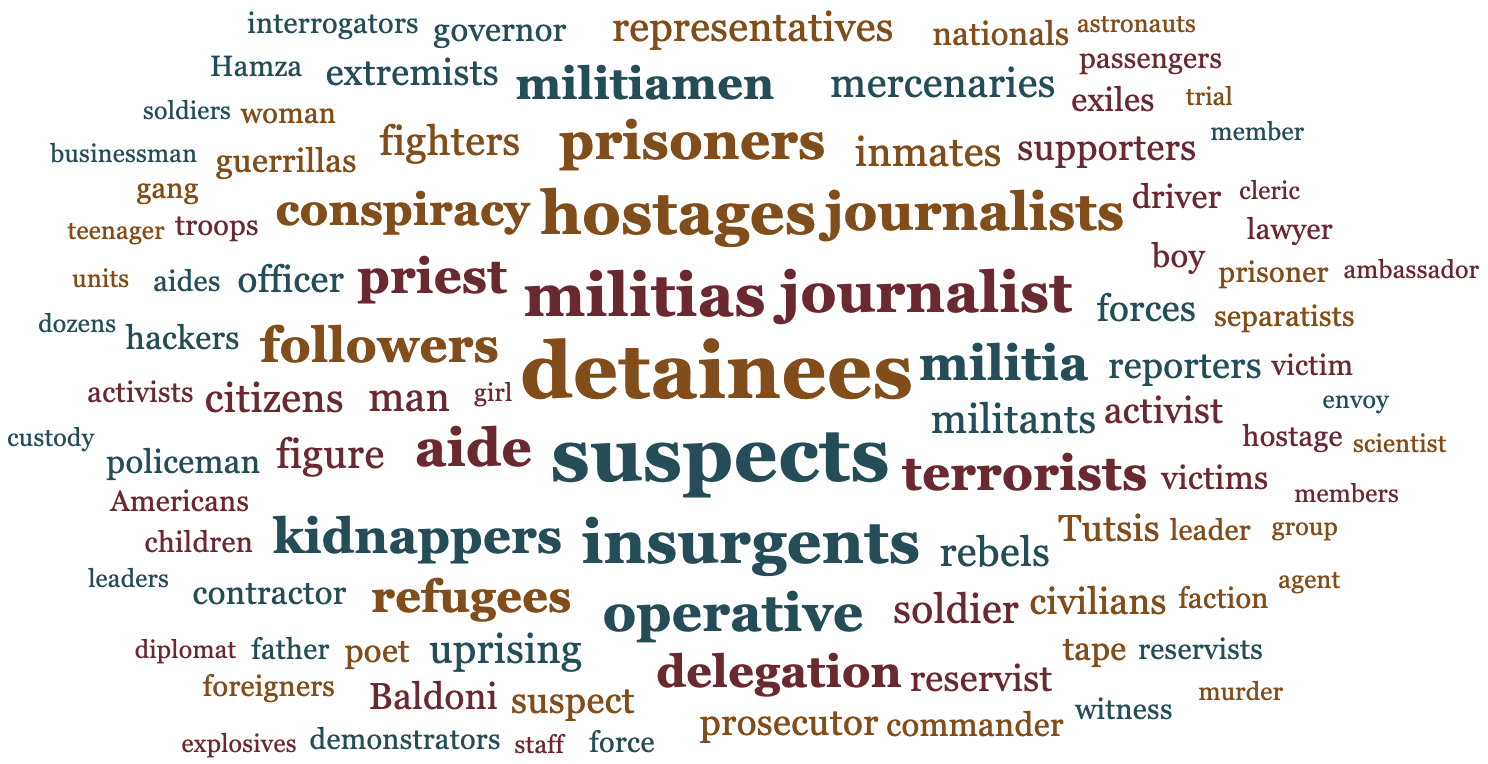} 
        \caption{Armed Conflict (World)}
        \label{fig:conflicts}
    \end{subfigure}
    \begin{subfigure}[b]{0.32\textwidth}
        \centering
        \includegraphics[width=\linewidth]{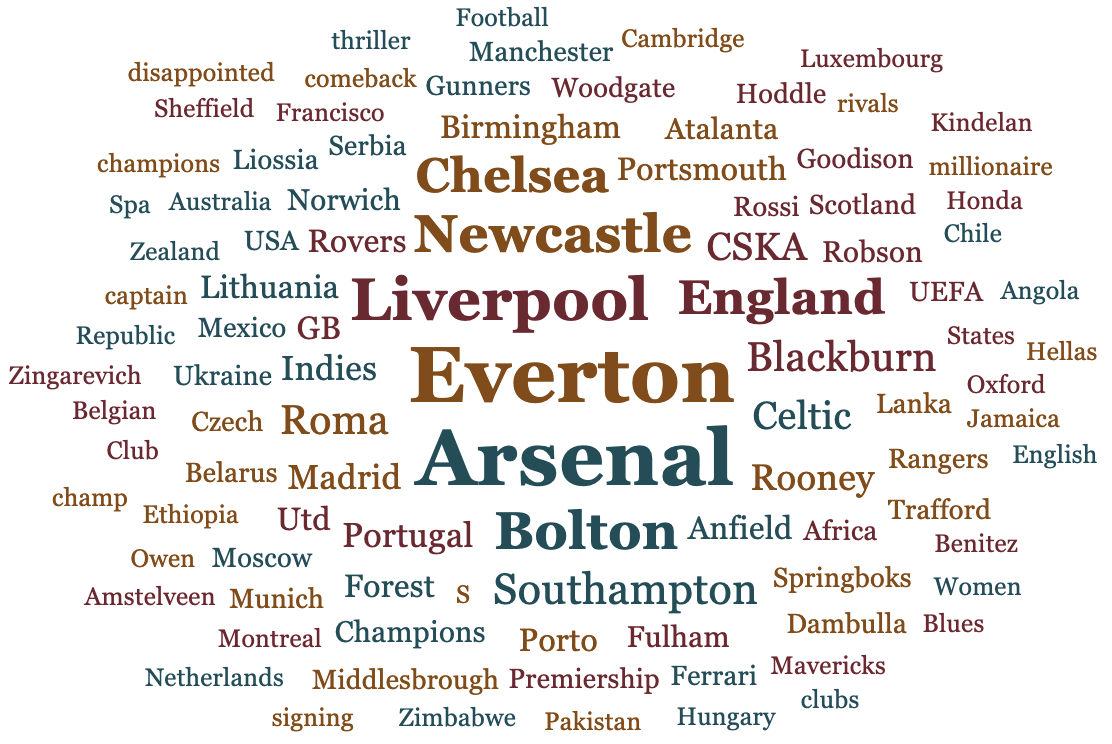} 
        \caption{Football (Sports)}
        \label{fig:football}
    \end{subfigure}
    \hfill
    \begin{subfigure}[b]{0.32\textwidth}
        \centering
        \includegraphics[width=\linewidth]{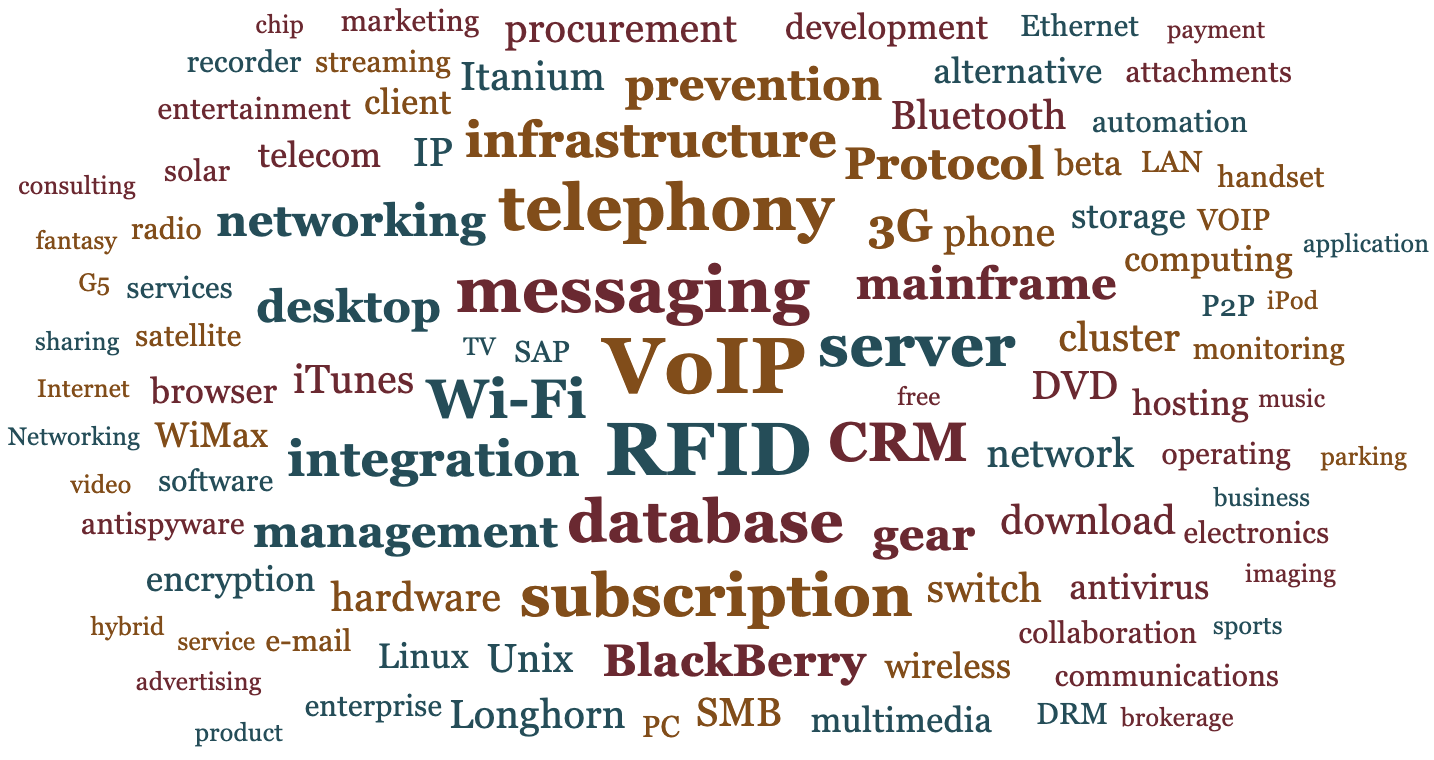} 
        \caption{Network Infrastructure (Sci/Tech)}
        \label{fig:networking}
    \end{subfigure}
    \hfill
    \begin{subfigure}[b]{0.32\textwidth}
        \centering
         \includegraphics[width=\linewidth]{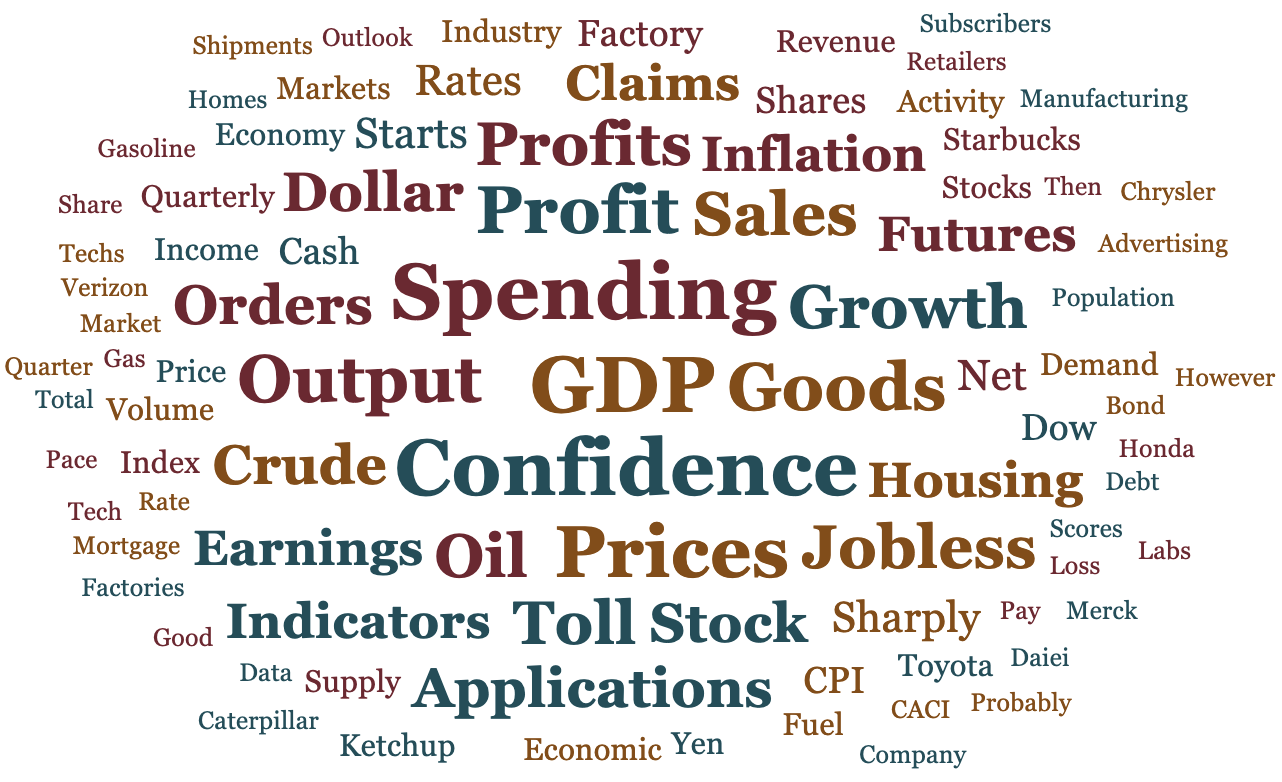} 
        \caption{Market Indicators (Business)}
        \label{fig:business}
    \end{subfigure}
    
    \caption{Examples of latent concepts identified in the Qwen model for the AG News classification task}
    \vspace{-8pt}
    \label{fig:wordclouds}
\end{figure*}



We analyze the latent concepts discovered by \gls{vqlc}. Each concept is visualized using a wordcloud constructed from the top-100 most frequent tokens. Figure~\ref{fig:wordclouds} presents examples of latent concepts learned by Qwen on the AG News dataset, with concepts drawn from each of the \textit{Sports}, \textit{Sci/Tech}, \textit{World}, and \textit{Business}. Within the \textit{Sports} category, Figure~\ref{fig:baseball} captures baseball players, with player names such as ``Sheffield'', ``Ramirez'', and ``Jeter''. Figure~\ref{fig:football} focuses on football, emphasizing team names such as ``Everton'', ``Arsenal'', and ``Chelsea''. For the \textit{Sci/Tech} category, Figure~\ref{fig:clinical} highlights clinical drugs and cancer treatment, with concept words such as ``clinical'', ``cancer'', and ``vaccine''. Figure~\ref{fig:networking} captures networking and telecom infrastructure, including terms such as ``VoIP'', ``Wi-Fi'', and ``networking''. Within the \textit{World} category, Figure~\ref{fig:conflicts} captures a conflict-related concept focused on insurgents, and militias, with words such as ``detainees'', ``militias'', and ``insurgents''. For the \textit{Business} category, Figure~\ref{fig:business} focuses on macroeconomics and market indicators, with words such as ``GDP'', ``Oil'' and ``Earnings''. 

We also include qualitative examples to check whether the discovered concepts capture task-related semantic information in Appendix~\ref{app:qualitative}. For example, Figure~\ref{fig:ag_qwen_729} shows a correct \textit{Business} prediction, where \gls{vqlc} retrieves a business concept focused on oil prices, GDP-related economic data, and treasury market reactions, while K-Means and \gls{lacoat} return broader economic clusters. Figure~\ref{fig:ag_qwen_207} shows an incorrect \textit{Business} prediction for a \textit{Science/Tech} instance. All methods reveal that the representation follows a business semantic direction, but \gls{vqlc} more directly captures corporate competition and consumer-hardware themes.

\subsection{Comparison with Sparse Autoencoders}
\Glspl{sae} expose sparse feature directions distributed across the model representation space, while \gls{vqlc} organizes representations into a set of task-level latent concepts. Because the two methods represent and organize semantic information differently, we treat \glspl{sae} as a complementary comparison rather than as another baseline. We compare them across three datasets using the Qwen model. Token representations from the last layer are passed through a pretrained \gls{sae}. \Gls{sae}-concept assigns tokens to feature indices, and forms a concept vector by averaging the hidden representations assigned to the same feature. In \gls{sae}-concept top1, each token is assigned to its maximally activated feature. Tokens sharing the same top-1 feature are treated as instances of a concept, and the corresponding averaged hidden representations are used as the ablated vector.

Table~\ref{tab:vqlc_sae} characterizes how each method's units behave on the task-decision representation. \Gls{sae} produces more features, but rarely activates them in the test set. \Gls{vqlc} generates fewer concepts, activates on a larger fraction of test examples, and produces greater confidence changes. Appendix~\ref{app:sae} provides additional results on Qwen and Gemma.

\section{Related Work}
Early interpretability methods focused on attributing input features to predictions, including \gls{ig}~\citep{Sundararajan2017}, SmoothGrad~\citep{smilkov2017smoothgrad}, SHAP~\citep{lundberg2017unified}, and LIME~\citep{ribeiro2016should}. Concept-based methods move beyond individual input features. TCAV \citep{kim2018interpretability} and CEBab \citep{abraham2022cebab} use human-defined concepts, while later works like ACE \citep{ghorbani_towards_2019} automatically discover concepts via clustering. More recent work extracts latent concepts directly from hidden representations to support higher-level semantic explanations~\citep{rajani2020explaining, dalvi2022discovering, jourdan2023cockatiel, zhao2024explaining, yu-etal-2024-latent, lam-2024, sharma2025analyzing}.


\Glspl{sae}, motivated by the superposition hypothesis, have become a central tool in mechanistic interpretability for decomposing representations into sparse feature directions~\citep{cunningham2023sparse, harle2024scar, templeton2024scaling, lan2024quantifying}. Unlike latent concept methods, \gls{sae} operate at the feature level and aim to disentangle the representation into monosemantic features rather than task-related concepts.
Therefore, we treat them as a complementary comparison rather than as a direct
baseline.


\Gls{vqvae}~\citep{Oord2017} learns discrete latent representations via a finite codebook, enabling a natural discretization of the representation space. Prior work has mainly used this idea for representation learning and compression~\citep{kaiser2018, guo2020, yu2021vector, bhardwaj-etal-2022-vector, huang2023}. \Gls{vqlc} instead uses vector quantization as a scalable mechanism for discovering latent concepts in \gls{llm} hidden states.

\section{Conclusion}
This work introduced \gls{vqlc}, a vector quantization-based framework for latent concept discovery. Across 12 dataset-model settings, \gls{vqlc} demonstrates a favorable balance between concept quality and scalability: it remains competitive with hierarchical and K-Means clustering on faithfulness, with the clearest gains on decoder-only \glspl{llm}, while requiring near-constant memory as the number of tokens grows. Compared with \gls{sae}, \gls{vqlc} offers a concept-level explanation that encodes more task-relevant information in the task-decision representation. Overall, these results position \gls{vqlc} as a practical and scalable approach for latent concept discovery in \glspl{llm}.


\section*{Limitations}
The \gls{vqlc} framework involves design choices that affect the learned concepts, including the codebook size, commitment weight, and dead-code recovery settings. These choices influence both reconstruction quality and concept granularity, and therefore require careful tuning. Moreover, the current study focuses on classification tasks. Since generative models rely on token-by-token generation and reasoning, extending latent concept-based explanation to analyze intermediate generation behavior is an essential direction for future work.

\bibliography{custom}

\newpage

\appendix

\section{Experiment Setup: Hardware}
\label{app:setup}
Our experiments were conducted on \gls{hpc} cluster equipped with NVIDIA H100 and L40 GPUs, and standard CPU resources. We performed all training and inference runs for \gls{vqlc} and K-Means on H100 GPUs.

\section{Dataset}
\label{app:dataset}

\paragraph{Data statistics}
We conduct experiments on three sequence classification tasks: ERASER Movie Reviews~\citep{eraser_sst} for sentiment classification task, Jigsaw Toxicity~\citep{cjadams_2017} for toxicity classification task, and AG News~\citep{gulli2005ag} for multi-class news topic classification. 

\begin{table}[!htbp]
\centering			
\small
    \begin{tabular}{l|c|c|c}					
    \toprule									
    \textbf{Benchmark}    & \textbf{Train} & \textbf{Dev} & \textbf{Tags}\\		
    \midrule
    ERASER Movie & 13878 & 856 & 2\\
    JIGSAW Toxicity & 9000 & 800 & 2\\
    AG News & 16000 & 1200 & 4\\
    \bottomrule
    \end{tabular}
\caption{Data statistics for the benchmarks used in the evaluation.}
\label{datasets_info}
\end{table}

\paragraph{Processing}
To keep the comparison with~\citet{yu-etal-2024-latent} aligned, we apply the same token filtering criteria, excluding those with frequencies lower than five, and randomly select 20 contextual occurrences of each token. We retain all occurrences for the representative classification tokens. For concept presentation and evaluations, we apply a lightweight post-processing step to concept token lists. We remove empty tokens, punctuation tokens, and common stopwords, while retaining digits and special tokens.

\section{Finetuning Performance of 12-layered pre-trained Models}
\label{app:models}

We finetuned two 12-layered pre-trained models: BERT-base-cased~\citep{devlin2018bert} and RoBERTa~\citep{liu2019roberta} with their standard data split. 

\begin{table}[h]
\centering	
\small
\setlength{\tabcolsep}{2.5pt}
    \begin{tabular}{l|c|c}									
    \toprule									
    \textbf{Benchmark} & \textbf{BERT} & \textbf{RoBERTa} \\		
    \midrule
    ERASER Movie & 93.74 & 95.98\\
    JIGSAW Toxicity & 91.30 & 91.66\\
    AG News & 94.88 & 95.18\\
    \bottomrule
    \end{tabular}
\caption{The fine-tuned performance of models across all benchmarks. Model: BERT, RoBERTa}
\label{app:table:finetune-stat}
\end{table}

\begin{table*}[t]
\centering
\small
\setlength{\tabcolsep}{2.5pt}
\begin{tabular}{lllcc}
\toprule
\textbf{Dataset} & \textbf{Model} & \textbf{Encoder} & \textbf{Confidence Change} & \textbf{\% of Prediction Change} \\
\midrule

\multirow{4}{*}{ERASER}
& \multirow{2}{*}{RoBERTa}
& Residual MLP & \textbf{0.4926 $\pm$ 0.0015} & \textbf{0.4560 $\pm$ 0.0089} \\
& & Linear       & 0.4869 $\pm$ 0.0013         & 0.4171 $\pm$ 0.0085 \\
\cmidrule(lr){2-5}
& \multirow{2}{*}{Qwen}
& Residual MLP & \textbf{0.0897 $\pm$ 0.0030} & \textbf{0.0829 $\pm$ 0.0011} \\
& & Linear       & 0.0782 $\pm$ 0.0045         & 0.0716 $\pm$ 0.0020 \\
\midrule

\multirow{4}{*}{AG News}
& \multirow{2}{*}{RoBERTa}
& Residual MLP & \textbf{0.7260 $\pm$ 0.0028} & \textbf{0.8750 $\pm$ 0.0029} \\
& & Linear       & 0.7168 $\pm$ 0.0011         & 0.8089 $\pm$ 0.0167 \\
\cmidrule(lr){2-5}
& \multirow{2}{*}{Qwen}
& Residual MLP & \textbf{0.1950 $\pm$ 0.0045} & \textbf{0.1960 $\pm$ 0.0045} \\
& & Linear       & 0.1906 $\pm$ 0.0033         & 0.1857 $\pm$ 0.0015 \\
\bottomrule
\end{tabular}
\caption{Ablation analysis of the nonlinear encoder branch layer. Each cell reports confidence change and the percentage of predictions changed after orthogonal projection. 
}
\label{tab:projection_results}
\end{table*}

\begin{table*}[h]
\centering
\small
\setlength{\tabcolsep}{2.5pt}
\begin{tabular}{lllcc}
\toprule
\textbf{Dataset} & \textbf{Model} & \textbf{Initialization} & \textbf{Confidence Change} & \textbf{\% of Prediction Change} \\
\midrule

\multirow{4}{*}{ERASER}
& \multirow{2}{*}{RoBERTa}
& Farthest-First & \textbf{0.4926 $\pm$ 0.0015} & \textbf{0.4560 $\pm$ 0.0089} \\
& & Random        & 0.4783 $\pm$ 0.0011         & 0.2932 $\pm$ 0.0229 \\
\cmidrule(lr){2-5}
& \multirow{2}{*}{Qwen}
& Farthest-First & \textbf{0.0897 $\pm$ 0.0037} & \textbf{0.0829 $\pm$ 0.0014} \\
& & Random        & 0.0777 $\pm$ 0.0046         & 0.0728 $\pm$ 0.0036 \\
\midrule

\multirow{4}{*}{AG News}
& \multirow{2}{*}{RoBERTa}
& Farthest-First & \textbf{0.7260 $\pm$ 0.0028} & \textbf{0.8750 $\pm$ 0.0029} \\
& & Random        & 0.7132 $\pm$ 0.0006         & 0.8008 $\pm$ 0.0029 \\
\cmidrule(lr){2-5}
& \multirow{2}{*}{Qwen}
& Farthest-First & 0.1950 $\pm$ 0.0045         & 0.1960 $\pm$ 0.0045 \\
& & Random        & \textbf{0.2264 $\pm$ 0.0008} & \textbf{0.2264 $\pm$ 0.0025} \\
\bottomrule

\end{tabular}
\caption{Comparison between Farthest-First and Random initialization. Each cell reports confidence change and the percentage of predictions changed after orthogonal projection. 
}
\label{tab:init}
\end{table*}

\section{Methodology: Additional Details}
\label{app:methodology}

\paragraph{Farthest-first initialization}
Let $\mathcal{Z} = \{z_1, \dots, z_M\}, \quad z_i \in \mathbb{R}^{d_c}$ denote the encoded token representations collected for initialization, and let $\tilde{\mathcal{Z}} = \{\tilde{z}_1, \dots, \tilde{z}_{M'}\}$ be the deduplicated candidates after rounded-value deduplication. Let $\bar{z}$ denote the mean of the deduplicated candidate pool. We choose the first vector as:
\begin{align}
    s_1 = \arg\max_{\tilde{z} \in \tilde{\mathcal{Z}}} \|\tilde{z} - \bar{z}\|_2^2.
\end{align}
This initialization favors a point that is well separated from the center of the candidates and provides a strong starting anchor in the encoded space. After selecting $t-1$ vectors, the next vector is chosen as the point whose distance to its nearest previously selected vector is maximal:
\begin{align}
    s_t = \arg\max_{\tilde{z} \in \tilde{\mathcal{Z}} \setminus \mathcal{S}_{t-1}} \min_{s \in \mathcal{S}_{t-1}} \|\tilde{z} - s\|_2^2,
\end{align}
where $\mathcal{S}_{t-1} = \{s_1, \dots, s_{t-1}\}$. We repeat this process until $K$ vectors are selected, and use them to initialize the codebook:
\begin{align}
    e_k \leftarrow s_k, \qquad k = 1, \dots, K.
\end{align}

\paragraph{Deadcode Recovery}
We maintain an inactivity counter for each code. At training step $t$, let 
\begin{align}
   u_j^{(t)} = \sum_i \mathds{1}\left[z_q(w_i) = e_j\right]
\end{align}
denote the number of assignments received by code $e_j$ in the current batch. We update the inactivity counter $d_j^{(t)}$ as 
\begin{align}
   d_j^{(t)} = 
   \begin{cases}
   d_j^{(t-1)} + 1, & \text{if } u_j^{(t)} \leq \delta, \\
   0, & \text{otherwise},
   \end{cases}
\end{align}
where $\delta$ is a small usage threshold. A code is considered eligible for recovery once its inactivity counter exceeds a patience threshold $T$. For each valid token, we define the assignment error as the squared Euclidean distance to its assigned code: 
\begin{align}
   a_i^{(t)} = \left\| z_e(w_i) - z_q(w_i) \right\|_2^2.
\end{align}
Recovered codes are reinitialized using encoder outputs with the largest assignment errors in the current batch.

\section{\gls{vqlc} hyperparameters}
\label{app:hyperparameters}
For the encoder, the nonlinear branch uses a hidden dimension of $128$. During vector quantization, the codebook size is set to $K = 400$, the commitment weight to $0.25$, and the \gls{ema} decay to $0.99$. We use dead-code recovery with a zero-assignment threshold, a patience of 100 training steps, and at most two code recoveries per step. Appendix~\ref{app:codebook_size} analyzes the codebook size choice.

\section{Ablation Study: Encoder}
\label{app:encoder}

We have an ablation study to evaluate the usefulness of the nonlinear encoder branch layer. We ablate the nonlinear encoder branch by replacing the default residual-\gls{mlp} encoder (\textbf{ResidualMLP}) with a linear projection encoder (\textbf{Linear}). We compare the faithfulness performance. Table~\ref{tab:projection_results} shows that the nonlinear encoder branch layer provides benefits across both encoder-based models and decoder-only models.






\section{Comparison of Codebook Initialization Methods}
\label{app:initialization}

We study the effect of codebook initialization by comparing the default farthest-first initialization against a random initialization strategy. In our method, the codebook is initialized from encoded training representations using the farthest-first mechanism in Section~\ref{sec:codebook_initializaton}. In random initialization, the codebook is initialized from a uniform distribution. Table~\ref{tab:init} indicates farthest first initialization is favored in most settings, which provides the more stable overall choice.

\section{Sensitivity Analysis: Codebook Size}
\label{app:codebook_size}

We have a sensitivity analysis of the codebook size on the AG News dataset using the RoBERTa model. Table~\ref{tab:codesize_ablation} shows that the codebook size of 400 achieves the best performance. In addition, 400 corresponds to the number of clusters used in the evaluation experiments of the \gls{lacoat} baseline~\citep{yu-etal-2024-latent}.

\begin{table}[h]
 \centering
 \small
 \resizebox{\columnwidth}{!}{%
 \begin{tabular}{lcc}
 \toprule
 \textbf{Setting} & \textbf{Confidence Change} & \textbf{\% of Prediction Change} \\
 \midrule
 200 & 0.720 $\pm$ 0.002 & 0.818 $\pm$ 0.010 \\
 400 (default) & \textbf{0.729 $\pm$ 0.003} & \textbf{0.879 $\pm$ 0.007} \\
 800 & 0.712 $\pm$ 0.001 & 0.789 $\pm$ 0.014 \\
 \bottomrule
 \end{tabular}
 }
 \caption{Codebook size sensitivity on the AG News/RoBERTa setting under faithfulness evaluation. Each cell reports confidence change and the percentage of predictions changed after orthogonal projection. \textbf{Higher is better ($\uparrow$)}. The best results are \textbf{bolded}.
 }
 \label{tab:codesize_ablation}
\end{table}

\section{Ablation Study: Dead-code Recovery}
\label{app:deadcode_recovery}

We have an ablation study to evaluate the usefulness of dead-code recovery. We compare the faithfulness performance with and without the dead-code recovery mechanism. Table~\ref{tab:dcr_ablation} shows that dead-code recovery mechanism provides better performance.

\begin{table}[h]
 \centering
 \small
 \resizebox{\columnwidth}{!}{%
 \begin{tabular}{lcc}
 \toprule
 \textbf{Setting} & \textbf{Confidence Change} & \textbf{\% of Prediction Change} \\
 \midrule
 \textbf{Enabled (default)} & \textbf{0.729 $\pm$ 0.003} & \textbf{0.879 $\pm$ 0.007} \\
 Disabled & 0.714 $\pm$ 0.011 & 0.834 $\pm$ 0.016 \\
 \bottomrule
 \end{tabular}
 }
 \caption{Dead-code recovery ablation on the AG News/RoBERTa setting under faithfulness evaluation. \textbf{Higher is better ($\uparrow$)}. The best results are \textbf{bolded}.}
 \label{tab:dcr_ablation}
\end{table}

\section{Sensitivity Analysis: Commitment Weight}
\label{app:commit}

\begin{table}[h]
 \centering
 \small
 \resizebox{\columnwidth}{!}{%
 \begin{tabular}{lcc}
 \toprule
 \textbf{Weight} & \textbf{Confidence Change} & \textbf{\% of Prediction Change} \\
 \midrule
 0.10 & 0.715 $\pm$ 0.003 & 0.806 $\pm$ 0.012 \\
 0.25 (default) & \textbf{0.729 $\pm$ 0.003} & \textbf{0.879 $\pm$ 0.007} \\
 0.50 & 0.717 $\pm$ 0.001 & 0.812 $\pm$ 0.008 \\
 \bottomrule
 \end{tabular}
 }

 \caption{Commitment weight sensitivity on the AG News-RoBERTa setting under faithfulness evaluation. \textbf{Higher is better ($\uparrow$)}. The best results are \textbf{bolded}.}
 \label{tab:commitment_ablation}
\end{table}

\begin{table*}[t]
 \centering
  \small
     \begin{tabular}{llcccc}
     \toprule
     \multicolumn{6}{l}{\textbf{\% of Prediction Change} ($\uparrow$)} \\
     \cmidrule(l){1-6}
     \textbf{Dataset} & \textbf{Method} & \textbf{RoBERTa} & \textbf{BERT} & \textbf{LLaMA} & \textbf{Qwen} \\
     \midrule
     \multirow{3}{*}{\textbf{AG News}}
     & \gls{vqlc}   & \textbf{0.875 $\pm$ 0.003} & 0.879 $\pm$ 0.003          & \textbf{0.354 $\pm$ 0.002} & 0.196 $\pm$ 0.004 \\
     & K-Means      & 0.827 $\pm$ 0.000          & 0.336 $\pm$ 0.000          & 0.350 $\pm$ 0.017          & \textbf{0.204 $\pm$ 0.000} \\
     & \gls{lacoat} & 0.829    & \textbf{0.881 }  & 0.350       & 0.190  \\
     \midrule
     \multirow{3}{*}{\textbf{ERASER}}
     & \gls{vqlc}   & 0.456 $\pm$ 0.011          & 0.480 $\pm$ 0.006          & \textbf{0.369 $\pm$ 0.003} & \textbf{0.083 $\pm$ 0.001} \\
     & K-Means      & \textbf{0.482 $\pm$ 0.000} & \textbf{0.519 $\pm$ 0.000} & 0.265 $\pm$ 0.004          & 0.061 $\pm$ 0.000 \\
     & \gls{lacoat} & 0.350         & 0.510          & 0.315       & 0.059 \\
     \midrule
     \multirow{3}{*}{\textbf{Jigsaw}}
     & \gls{vqlc}   & 0.436 $\pm$ 0.011          & 0.498 $\pm$ 0.023          & \textbf{0.136 $\pm$ 0.006} & \textbf{0.158 $\pm$ 0.016} \\
     & K-Means      & 0.445 $\pm$ 0.000          & \textbf{0.543 $\pm$ 0.000} & 0.036 $\pm$ 0.009          & 0.150 $\pm$ 0.002 \\
     & \gls{lacoat} & \textbf{0.477} & 0.506        & 0.058          & 0.157 \\
     \bottomrule
     \end{tabular}
 \caption{Additional faithfulness results measured by percentage of predictions changed after projection-based concept ablation. \textbf{Higher is better ($\uparrow$)}. The best results are \textbf{bolded}.}
 \label{tab:faithfulness_prediction}
\end{table*}

\begin{table}[h]
\centering
\small
\begin{tabular}{llc}
\toprule
\textbf{Dataset} & \textbf{Method} & \textbf{BERT} \\
\midrule
\multirow{3}{*}{\textbf{AG News}}
& \gls{vqlc}   & \textbf{0.749 $\pm$ 0.004} \\
& K-Means      & 0.504 $\pm$ 0.000 \\
& \gls{lacoat} & 0.748  \\
\midrule
\multirow{3}{*}{\textbf{ERASER}}
& \gls{vqlc}   & 0.495 $\pm$ 0.005 \\
& K-Means      & \textbf{0.508 $\pm$ 0.000} \\
& \gls{lacoat} & 0.502 \\
\midrule
\multirow{3}{*}{\textbf{Jigsaw}}
& \gls{vqlc}   & 0.479 $\pm$ 0.001 \\
& K-Means      & \textbf{0.482 $\pm$ 0.000} \\
& \gls{lacoat} & 0.480 \\
\bottomrule
\end{tabular}
\caption{Additional faithfulness results for BERT, measured by confidence change after projection-based concept ablation. \textbf{Higher is better ($\uparrow$)}. The best results are \textbf{bolded}.}
\label{tab:faithfulness_bert}
\end{table}

The commonly recommended commitment weight in prior literature is $0.25$. In addition, we conduct a sensitivity analysis of different commitment weights on the AG News dataset using the RoBERTa model. We evaluate how varying the weight affects the faithfulness performance. Table~\ref{tab:commitment_ablation} demonstrates that a commitment weight of $0.25$ achieves the best overall performance. It has the highest confidence change and prediction percentage change. These results indicate that $0.25$ is the most stable choice.

\section{Faithfulness Evaluation}
\label{app:faithfulness}

\paragraph{Orthogonal Projection for Concept Ablation}

To measure whether an assigned latent concept encodes a direction that the underlying model relies on for prediction, we remove the corresponding latent concept direction from a sentence representation via orthogonal projection. Let $v_j$ denote the concept vector, and let $h_{w_i}^{(\ell)}$ denote the sentence representation at layer $\ell$. 

The projection of $h_{w_i}^{(\ell)}$ onto $v_j$ is defined as:
\begin{align}
\operatorname{proj}_{v_j}\!\left(h_{w_i}^{(\ell)}\right)
&= 
\frac{h_{w_i}^{(\ell)} \cdot v_j}{\lVert v_j \rVert^{2}} \, v_j 
\end{align}
This projection isolates the component of the representation that aligns with the concept direction.

We then remove this concept direction by subtracting the projection from the original representation:
\begin{align}
h_{w_i,\perp}^{(\ell)}
&=
h_{w_i}^{(\ell)}
- \operatorname{proj}_{v_j}\!\left(h_{w_i}^{(\ell)}\right)
\end{align}

The resulting representation $h_{w_i,\perp}^{(\ell)}$ preserves information of the original representation except for the contribution along the concept direction.



\paragraph{Additional Faithfulness Results}
Table~\ref{tab:faithfulness_bert} and Table~\ref{tab:faithfulness_prediction}
report additional BERT confidence change and prediction change of all dataset-model settings. For the BERT model, \gls{vqlc} and \gls{lacoat} remain close on all three datasets. For prediction change metrics, the overall pattern is consistent with confidence change results: decoder-only settings remain the advantages for \gls{vqlc}. \Gls{vqlc} has comparable performance for encoder-based settings.

\section{\glspl{llm}-Based Evaluation}
\label{app:llm}

\paragraph{Average Ranking Formula}
Let $S$ denote the set of evaluation samples, $A$ the set of method approaches, and $E$ the set of the \gls{llm} judges. For each sample $i \in S$, each \gls{llm} judge $e \in E$ assigns a ranking value $r_{i,a}^{(e)} \in \{1,2,3\}$ for each method $a \in A$, where a lower rank indicates a better explanation.

For each sample $i$ and method $a$, we have an aggregated rank value $\hat{r}_{i,a}$  by majority vote across the evaluators. A valid majority voting rank is defined only when at least two of the three evaluators assign the same rank value to that method. Let $\hat{S} \subseteq S$ denote the set of resolved samples. The average rank of method $a$ is then computed as:
\begin{align}
    \text{AvgRank}(a) = \frac{1}{|\hat{S}|}\sum_{i \in \hat{S}} \hat{r}_{i,a},
\end{align}
where $|\hat{S}|$ denotes the number of resolved samples. A lower average rank value indicates that the method is preferred more often by the \gls{llm} evaluators.

\paragraph{Additional \glspl{llm}-based Evaluation Results}
Table~\ref{tab:judge_results_bert} shows additional \glspl{llm}-based evaluation results for BERT across all datasets. \Gls{vqlc} achieves the best results in all settings. In additional, agreement scores are generally moderate to strong.

\begin{table}[ht]
\centering
\small

    \resizebox{\columnwidth}{!}{%
    \begin{tabular}{llcc}
    \toprule
    \textbf{Dataset} & \textbf{Method} & \textbf{Avg. Rank} ($\downarrow$) & \textbf{Kendall's $W$} ($\uparrow$) \\
    \midrule
    \multirow{3}{*}{\textbf{Jigsaw}}
    & \gls{vqlc}   & \textbf{1.721 $\pm$ 0.167} & \multirow{3}{*}{0.797 $\pm$ 0.011} \\
    & K-Means      & 2.483 $\pm$ 0.102          & \\
    & \gls{lacoat} & 1.751 $\pm$ 0.122          & \\
    \midrule
    \multirow{3}{*}{\textbf{ERASER}}
    & \gls{vqlc}   & \textbf{1.506 $\pm$ 0.118} & \multirow{3}{*}{0.804 $\pm$ 0.025} \\
    & K-Means      & 1.624 $\pm$ 0.037          & \\
    & \gls{lacoat} & 2.870 $\pm$ 0.038          & \\
    \midrule
    \multirow{3}{*}{\textbf{AG News}}
    & \gls{vqlc}   & \textbf{1.850 $\pm$ 0.083} & \multirow{3}{*}{0.726 $\pm$ 0.079} \\
    & K-Means      & 2.008 $\pm$ 0.027          & \\
    & \gls{lacoat} & 2.032 $\pm$ 0.037          & \\
        \bottomrule
        \end{tabular}
    }

\caption{Additional \glspl{llm}-based evaluation results for BERT. Average rank measures evaluator preference, and Kendall's $W$ measures inter-\glspl{llm} agreement over the three-method rankings}.
\label{tab:judge_results_bert}
\end{table}

\paragraph{Prompt Template}
We use the following prompt template for each \gls{llm} judge. For each sample, the order of the three candidate explanations is randomly shuffled to reduce position bias. The prompt always includes the input text, the model prediction, and three candidate concept-based explanations. Concepts are presented in one of two formats. If more than half of a concept consists of special tokens such as \texttt{[CLS]}
, we randomly sample five such tokens and provide their original sentences. Otherwise, we provide up to ten frequent tokens from the concept to fit the API context limit. Thus, concept content is presented in one of these two forms: (i) five sample sentences for special token dominated concepts, or (ii) a list of up to ten tokens.

 \begin{promptbox}
  You are an expert judge of local concept-based explanations.\par
  \par
  Your task is to rank candidate explanations for why the model made its prediction for a single input.\par
  \par
  Input Text: [INPUT\_TEXT]\par
  Model Prediction: [MODEL\_PREDICTION]\par
  \par
  Please evaluate the candidate explanations using the following principles:\par
  1. The best explanation should identify the most important semantic reason for why the model made its prediction.\par
  2. Prefer explanations whose concept content matches the specific topic, event, entity, or semantic pattern in the input.\par
  3. Do not reward generic topical overlap if another explanation is more specific and directly relevant to the prediction.\par
  4. When concepts are weak, noisy, or only loosely related to the prediction, rank them lower.\par
  \par
  Candidate Explanations:\par
  \par
  Explanation 1:\par
  Name: [METHOD\_NAME\_1]\par
  Concept Content: [CONCEPT\_CONTENT\_1]\par
  \par
  Explanation 2:\par
  Name: [METHOD\_NAME\_2]\par
  Concept Content: [CONCEPT\_CONTENT\_2]\par
  \par
  Explanation 3:\par
  Name: [METHOD\_NAME\_3]\par
  Concept Content: [CONCEPT\_CONTENT\_3]\par
  \par
  Return a JSON object with one field "ranking", mapping each explanation name to a rank from 1 to 3, where 1 is best. Ties are allowed.\par
  Also include a short field "reason" explaining the ranking.
\end{promptbox}

\section{Comparison with \Gls{sae}}
\label{app:sae}

Table~\ref{tab:vqlc_sae_pct} and Table\ref{tab:sae_feature_topk} show the full comparison results between \gls{vqlc} and \gls{sae} on the Qwen model. For \gls{sae}-concept, we use the average vector of the hidden representations assigned to the same feature. In the top1 setting, each token is assigned to its most highly activated feature. In the top5 setting, each token is assigned to its unordered top5 \gls{sae} feature set. This makes the \gls{sae}-concept comparable to \gls{vqlc}, where each explanation is represented by the average vector over its assigned token representations. 

\Gls{sae}-feature instead directly ablates the active \gls{sae} features by setting the selected activations to zero. The edited latent is then decoded back into hidden presentations and forward pass to measure the resulting performance effect. In general, \glspl{sae} produces more features, but their top1 and top5 features are activated in fewer test examples. \Gls{vqlc} use a smaller codebook and is activated on most inputs. \Gls{vqlc} concepts encode more task-relevant information used by the model for prediction in the Qwen setting.

\begin{table}[t]
\centering
\setlength{\tabcolsep}{2.5pt}
\resizebox{\columnwidth}{!}{%
\begin{tabular}{llcccc}
\toprule
\textbf{Dataset} & \textbf{Method} & \textbf{Confidence Change} & \textbf{\% Pred.\ Change} & \textbf{\# Concepts} & \textbf{Active Rate} \\
\midrule
\multirow{3}{*}{\textbf{AG News}}
& \gls{vqlc} & \textbf{0.4444} & \textbf{0.6045} & 399 & \textbf{0.8897} \\
& SAE-concept top1 & 0.3379 & 0.3878 & 1,834 & 0.3751 \\
& SAE-concept top5 & 0.1668 & 0.2943 & 176,251 & 0.0868 \\
\midrule
\multirow{3}{*}{\textbf{Jigsaw}}
& \gls{vqlc} & \textbf{0.3035} & \textbf{0.5676} & 398 & \textbf{0.7789} \\
& SAE-concept top1 & 0.0930 & 0.1071 & 918 & 0.1993 \\
& SAE-concept top5 & 0.1830 & 0.1154 & 62,149 & 0.1007 \\
\midrule
\multirow{3}{*}{\textbf{ERASER}}
& \gls{vqlc} & \textbf{0.2960} & \textbf{0.4417} & 398 & \textbf{0.7387} \\
& SAE-concept top1 & 0.0579 & 0.0206 & 601 & 0.2163 \\
& SAE-concept top5 & 0.1131 & 0.0874 & 42,572 & 0.1096 \\
\bottomrule
\end{tabular}
}
\caption{Full comparison between \gls{vqlc}, SAE-concept top1, and SAE-concept top5 on the Qwen model. SAE-concept first assigns tokens to \gls{sae} feature indices and then uses the average hidden state of the assigned tokens as the concept direction. We report faithfulness results, concept inventory size, and active rate.}
\label{tab:vqlc_sae_pct}
\end{table}

\begin{table}[h]
\centering
\setlength{\tabcolsep}{4pt}
\resizebox{\columnwidth}{!}{%
\begin{tabular}{llcc}
\toprule
\textbf{Dataset} & \textbf{Method} & \textbf{Confidence Change} & \textbf{\% Pred.\ Change} \\
\midrule
\multirow{2}{*}{\textbf{AG News}}
& SAE-feature top1 & 0.0300 & 0.0060 \\
& SAE-feature top5 & \textbf{0.2066} & \textbf{0.2089} \\
\midrule
\multirow{2}{*}{\textbf{Jigsaw}}
& SAE-feature top1 & 0.1027 & 0.0153 \\
& SAE-feature top5 & \textbf{0.3274} & \textbf{0.2844} \\
\midrule
\multirow{2}{*}{\textbf{ERASER}}
& SAE-feature top1 & 0.0433 & \textbf{0.0328} \\
& SAE-feature top5 & \textbf{0.0623} & 0.0291 \\
\bottomrule
\end{tabular}
}
\caption{Additional faithfulness results for SAE-feature top1 and SAE-feature top5 on the Qwen model. SAE-feature directly uses \gls{sae} feature directions, without averaging assigned token representations.}
\label{tab:sae_feature_topk}
\end{table}


We also compare \gls{vqlc} with \gls{sae} in \texttt{Gemma-3-4b-IT} at the last layer (see Table~\ref{tab:gemma_vqlc_sae} and Table~\ref{tab:gemma_sae_feature_topk}). These results broadly support the same pattern as the Qwen comparison. \Gls{vqlc} gives a much smaller concept inventory and a higher active rate. For \gls{sae}-concept, ERASER movie is the only exception where \gls{sae}-concept top5 gives a slightly higher performance change in faithfulness, but it has large concept inventory and a much lower active rate. For the \gls{sae}-feature, it is stronger in AG News. This shows that individual \gls{sae} feature can perturb output in some cases, but they do not provide the same consistently active concept as \gls{vqlc}.

\begin{table}[h]
\centering
\setlength{\tabcolsep}{2.5pt}
\resizebox{\columnwidth}{!}{%
\begin{tabular}{llcccc}
\toprule
\textbf{Dataset} & \textbf{Method} & \textbf{Confidence Change} & \textbf{\% Pred.\ Change} & \textbf{\# Concepts} & \textbf{Active Rate} \\
\midrule
\multirow{3}{*}{\textbf{AG News}}
& \gls{vqlc} & \textbf{0.1803} & \textbf{0.1758} & 397 & \textbf{0.8060} \\
& SAE-concept top1 & 0.0209 & 0.0192 & 2,788 & 0.4706 \\
& SAE-concept top5 & 0.0478 & 0.0427 & 420,462 & 0.0541 \\
\midrule
\multirow{3}{*}{\textbf{Jigsaw}}
& \gls{vqlc} & \textbf{0.1296} & \textbf{0.1224} & 398 & \textbf{0.8040} \\
& SAE-concept top1 & 0.0438 & 0.0383 & 2,346 & 0.3176 \\
& SAE-concept top5 & 0.0739 & 0.0744 & 280,026 & 0.0421 \\
\midrule
\multirow{3}{*}{\textbf{ERASER}}
& \gls{vqlc} & 0.0292 & 0.0269 & 399 & \textbf{0.7519} \\
& SAE-concept top1 & 0.0002 & 0.0000 & 1,583 & 0.3797 \\
& SAE-concept top5 & \textbf{0.0312} & \textbf{0.0292} & 170,980 & 0.0487 \\
\bottomrule
\end{tabular}
}
\caption{Full comparison between \gls{vqlc}, SAE-concept top1, and SAE-concept top5 on the Gemma model.}
\label{tab:gemma_vqlc_sae}
\end{table}

\begin{table}[h]
\centering
\setlength{\tabcolsep}{4pt}
\resizebox{\columnwidth}{!}{%
\begin{tabular}{llcc}
\toprule
\textbf{Dataset} & \textbf{Method} & \textbf{Confidence Change} & \textbf{\% Pred.\ Change} \\
\midrule
\multirow{2}{*}{\textbf{AG News}}
& SAE-feature top1 & 0.1974 & 0.1817 \\
& SAE-feature top5 & \textbf{0.2290} & \textbf{0.1933} \\
\midrule
\multirow{2}{*}{\textbf{Jigsaw}}
& SAE-feature top1 & \textbf{0.0392} & \textbf{0.0357} \\
& SAE-feature top5 & 0.0244 & 0.0167 \\
\midrule
\multirow{2}{*}{\textbf{ERASER}}
& SAE-feature top1 & 0.0123 & 0.0129 \\
& SAE-feature top5 & \textbf{0.0129} & \textbf{0.0140} \\
\bottomrule
\end{tabular}
}
\caption{Additional faithfulness results for SAE-feature top1 and SAE-feature top5 on the Gemma model.}
\label{tab:gemma_sae_feature_topk}
\end{table}

\section{Examples of Qualitative Evaluation}
\label{app:qualitative}

\paragraph{Comparing Latent Concept Methods}

\begin{figure*}[ht]
      \centering
      \includegraphics[width=\textwidth]{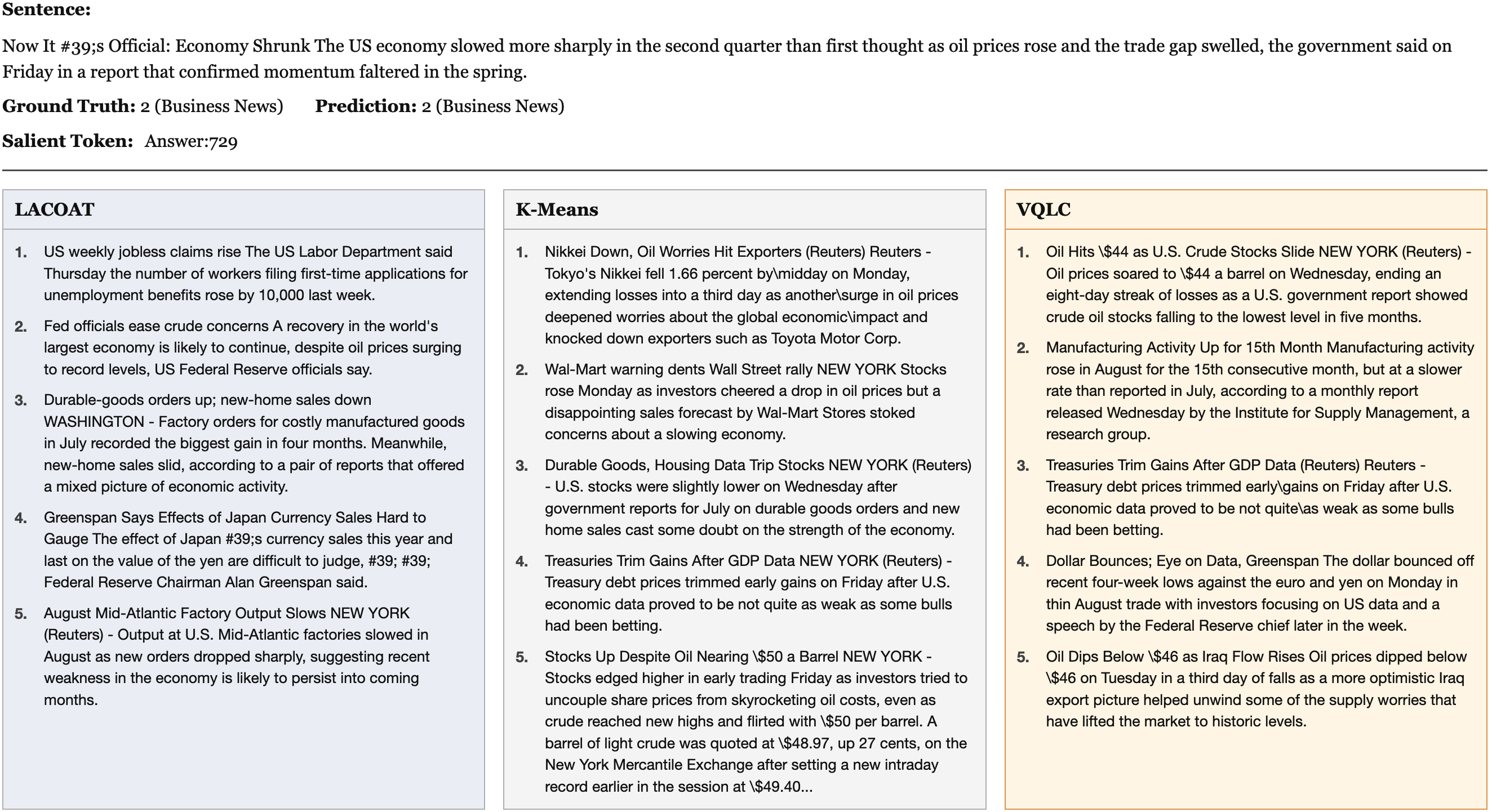}
      \caption{Correct prediction example from AG Qwen.}
      \label{fig:ag_qwen_729}
\end{figure*}

Figure~\ref{fig:ag_qwen_729} shows a correct prediction example, where the sample sentence discusses economic contraction, rising oil prices, and a widening grade gap, and the model correctly predicts the \textit{Business News} label. The concept identified by \gls{vqlc} is the most closely aligned with the sample sentence. The content retrieved directly focuses on oil prices, GDP-related economic data, and treasury market reactions. K-Means also captures economically related content, but the concept is broader, mixing oil-price movements with general stock-market reactions. \Gls{lacoat} discovers generic macroeconomic indicators such as jobless claims, factory output, and durable-goods orders, which are less directly related to the oil prices and economic slowdown emphasized in the sample sentence. Overall, \gls{vqlc} provides a more precise latent concept to explain the model predictions for this sample.

Figure~\ref{fig:ag_qwen_207} in the Appendix provides an example where the model incorrectly predicts \textit{Business News} label for a \textit{Science/Technology News}. The sample sentence concerns PC customer satisfaction and hardware support. All three methods reveal that the model relies on a business-oriented semantic direction for this prediction. K-Means finds the concept is centered more on specific contracts, acquisitions, and product launches. \Gls{lacoat} provides the concept related to the broader enterprise IT and branding. The concept identified by \gls{vqlc} focuses on business-oriented technology themes such as corporate competition and consumer hardware, which more closely match the semantic content of the input. It helps to understand why the model incorrectly predicts \textit{Business News} instead of \textit{Science/Technology News} for this sample explicitly. 

\begin{figure*}[h]
      \centering
      \includegraphics[width=\textwidth]{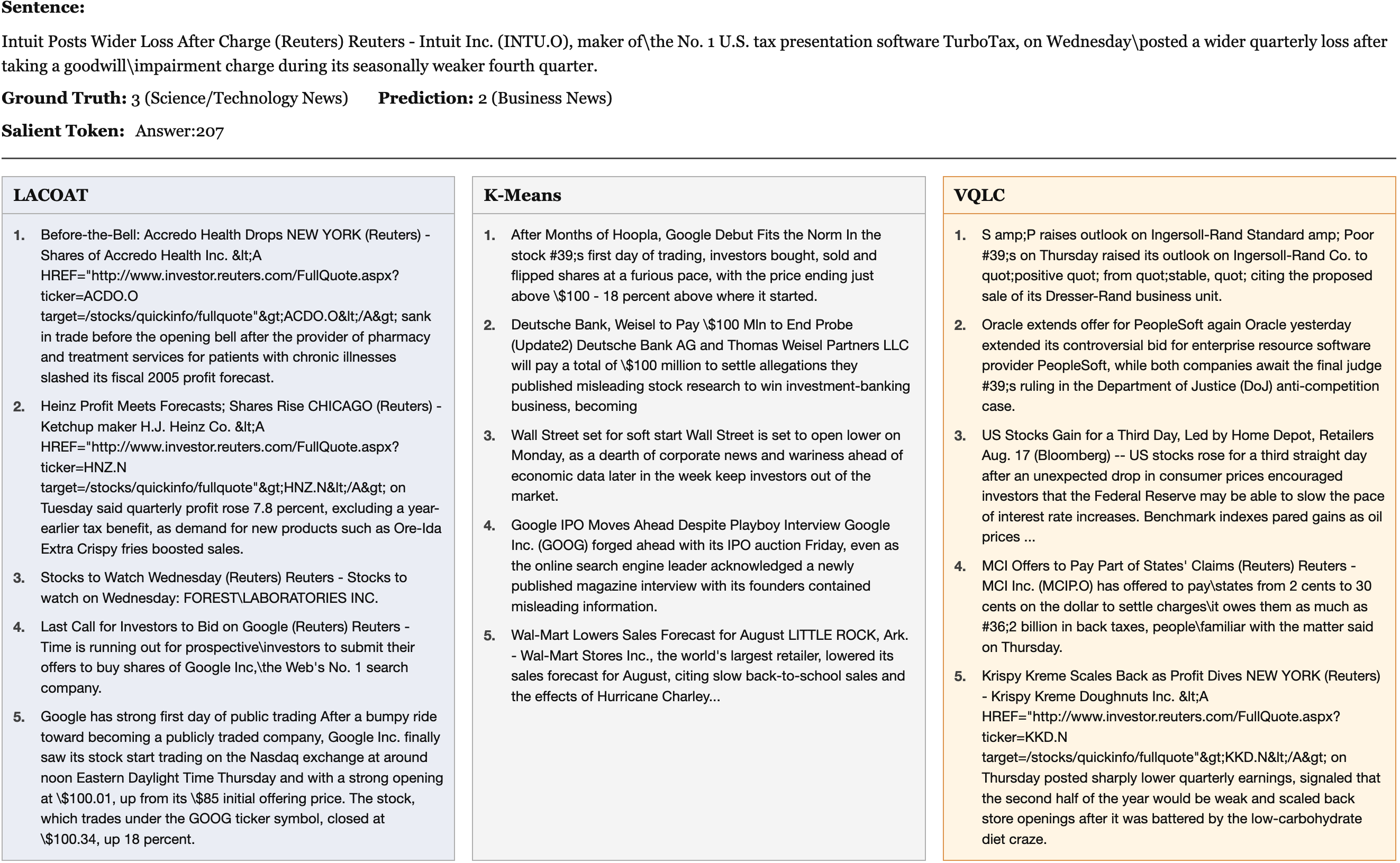}
      \caption{Incorrect prediction example from AG Qwen.}
      \label{fig:ag_qwen_207}
\end{figure*}

\begin{figure*}[h]
      \centering
      \includegraphics[width=\textwidth]{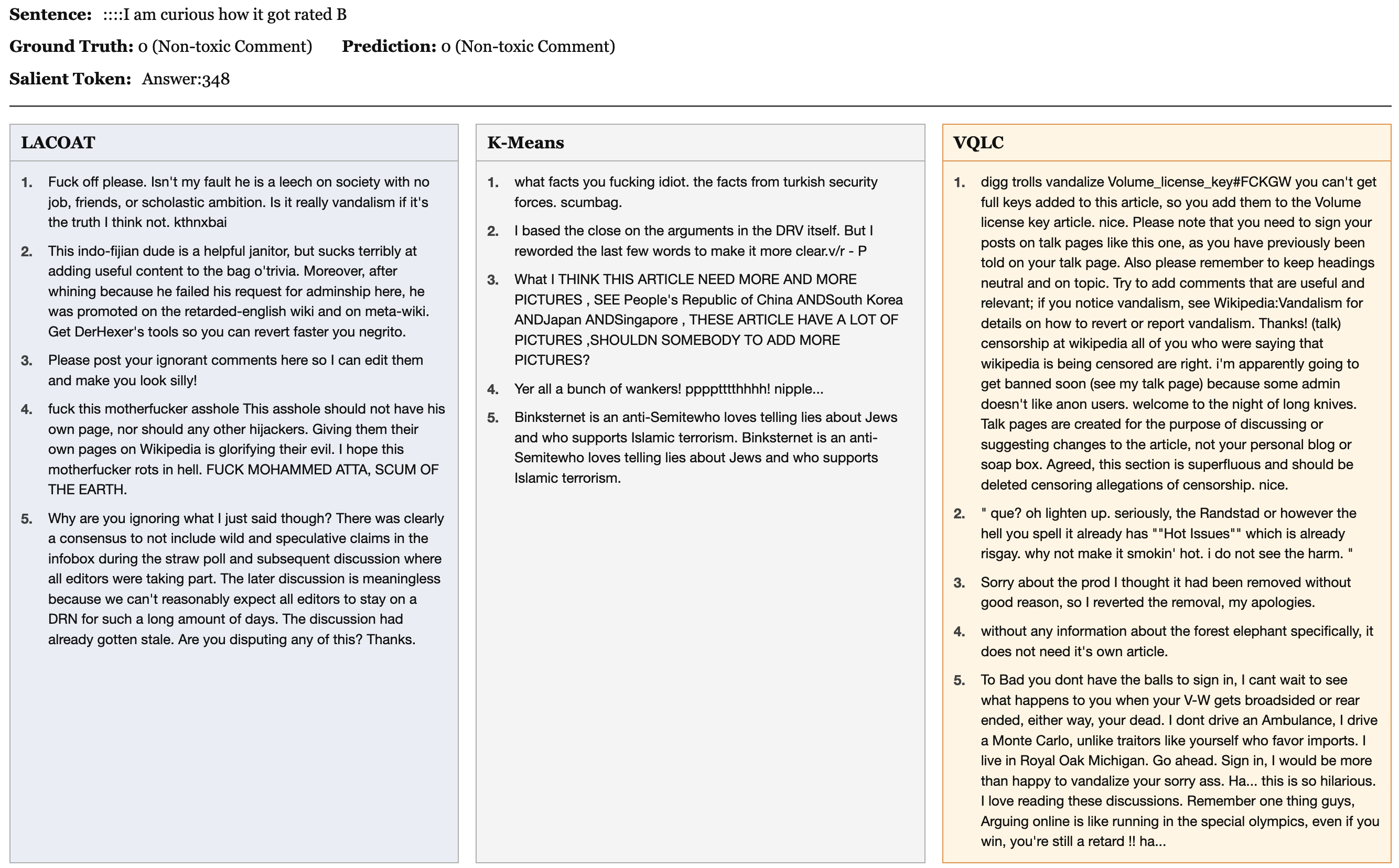}
      \caption{Correct prediction example from Jigsaw Qwen.}
      \label{fig:jigsaw_qwen_348}
\end{figure*}

Figure~\ref{fig:jigsaw_qwen_348} shows a correct prediction example for the Jigsaw dataset using the Qwen model. The concepts of both \gls{lacoat} and K-Means mainly contains toxic sentences. In contrast, most sentences in the concept of \gls{vqlc} are non-toxic. \Gls{vqlc} captures the discourse behaviors and non-toxic semantic, making it more suitable for interpreting model prediction. 

\begin{figure*}[h]
      \centering
      \includegraphics[width=\textwidth]{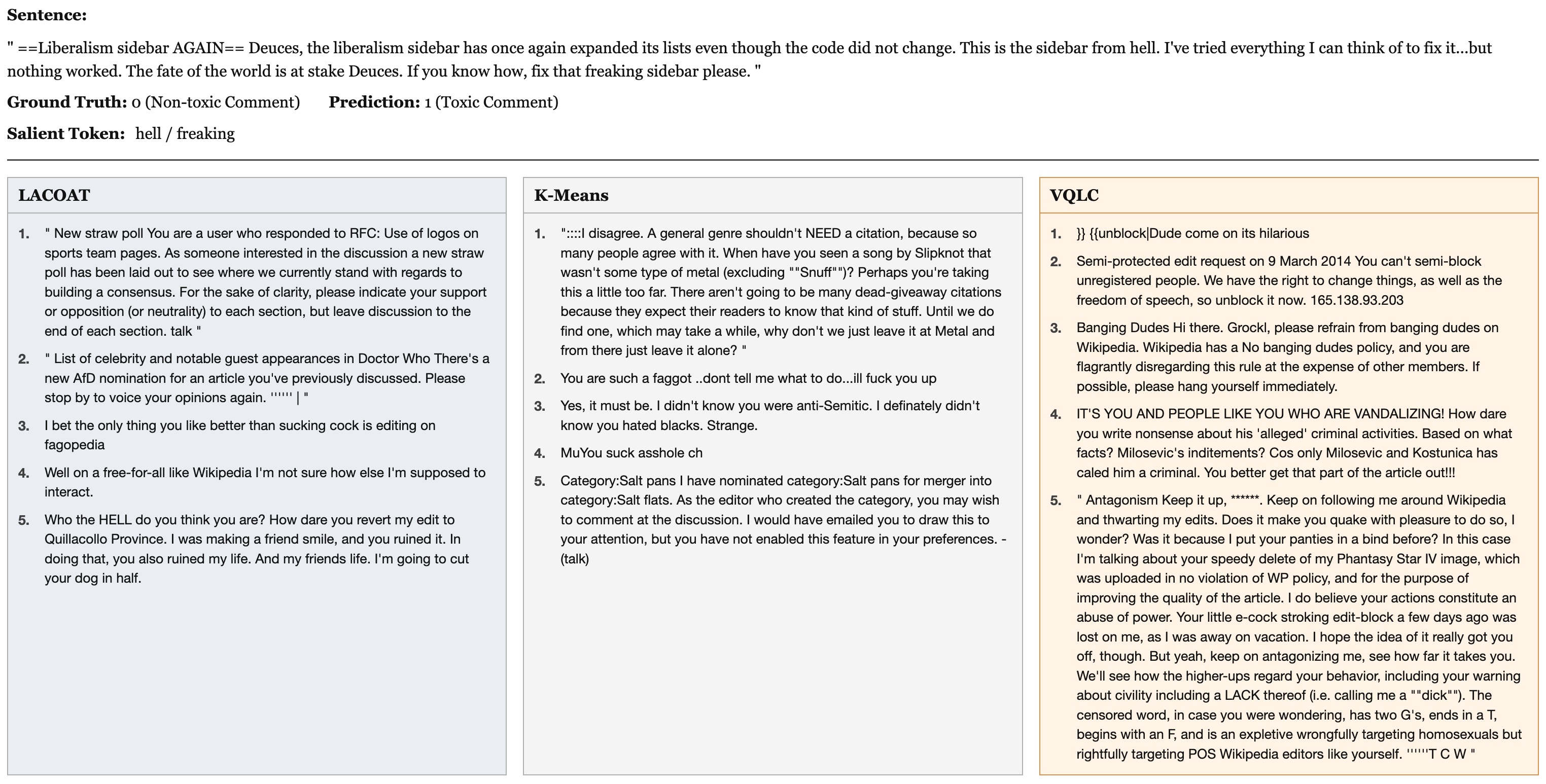}
      \caption{Incorrect prediction example from Jigsaw Qwen.}
      \label{fig:jigsaw_qwen_228}
\end{figure*}

Figure~\ref{fig:jigsaw_qwen_228} displays a case where a non-toxic sentence is misclassified as a toxic label. Although all three methods provide mixed concepts that contain both benign and aggressive examples, the underlying representation region itself is semantically ambiguous. K-Means associate emotional expression with strongly toxic lexical samples. \Gls{lacoat} groups together a broad range of conflict discussion. In contrast, \gls{vqlc} produces a concept focused on the semantics of argumentative interaction.

\begin{figure*}[th]
    \centering

    \begin{subfigure}[b]{0.32\textwidth}
        \centering
        \includegraphics[width=\linewidth]{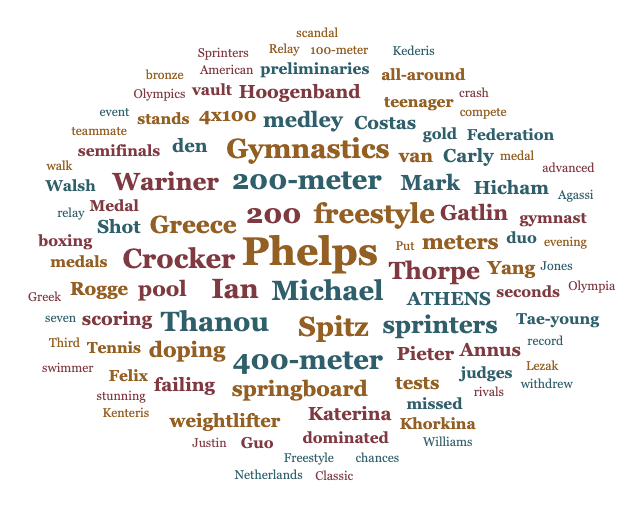} 
        \caption{Olympic Sports (Sports)}
        \label{fig:olympic}
    \end{subfigure}
    \hfill 
    \begin{subfigure}[b]{0.32\textwidth}
        \centering
        \includegraphics[width=\linewidth]{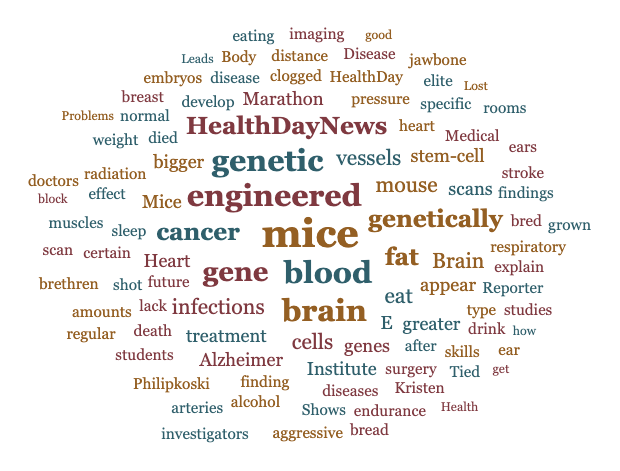} 
        \caption{Biomedical Research (Sci/Tech)}
        \label{fig:biomedical}
    \end{subfigure}
    \hfill
    \begin{subfigure}[b]{0.32\textwidth}
        \centering
        \includegraphics[width=\linewidth]{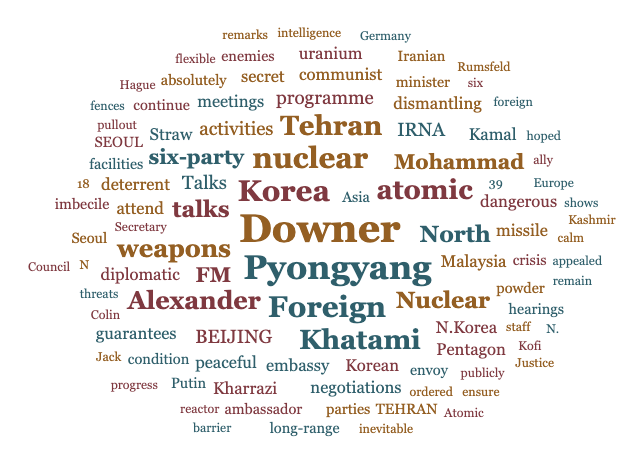} 
        \caption{Geopolitics (World)}
        \label{fig:nuclear}
    \end{subfigure}
    \begin{subfigure}[b]{0.32\textwidth}
        \centering
        \includegraphics[width=\linewidth]{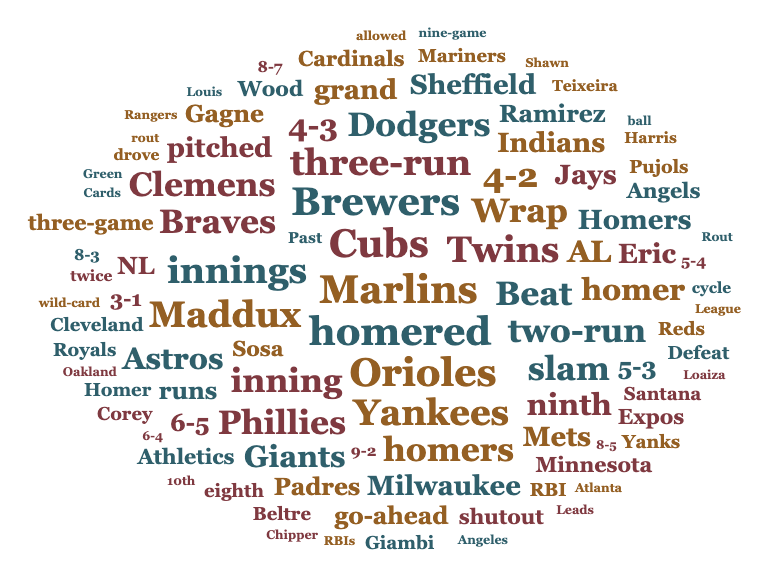} 
        \caption{Baseball (Sports)}
        \label{fig:baseball_roberta}
    \end{subfigure}
    \hfill
    \begin{subfigure}[b]{0.32\textwidth}
        \centering
        \includegraphics[width=\linewidth]{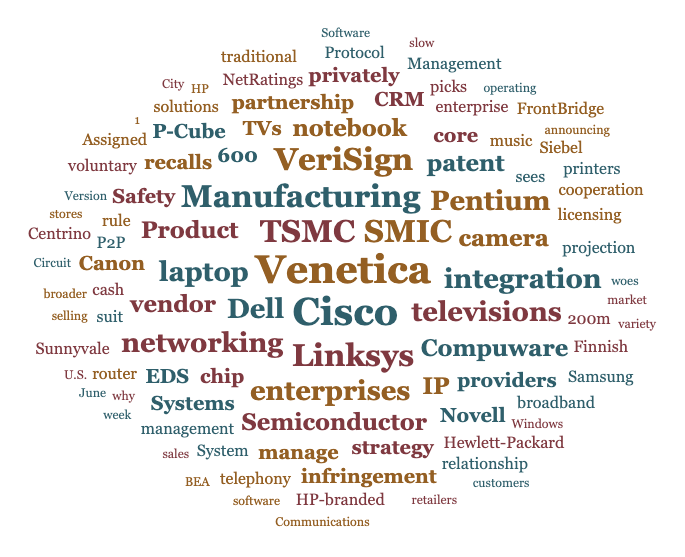} 
        \caption{Technology (Sci/Tech)}
        \label{fig:computer}
    \end{subfigure}
    \hfill
    \begin{subfigure}[b]{0.32\textwidth}
        \centering
        \includegraphics[width=\linewidth]{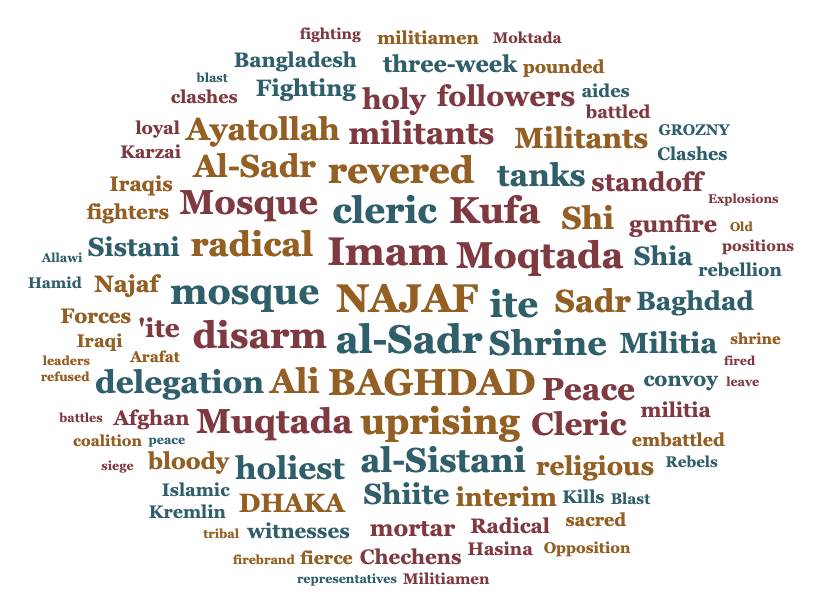} 
        \caption{Shiite Uprising in Iraq (World)}
        \label{fig:Iraq}
    \end{subfigure}
    
    \caption{Examples of latent concepts identified in the RoBERTa model for the AG News classification task}
    \vspace{-8pt}
    \label{fig:wordclouds_roberta}
\end{figure*}

\paragraph{Discovered Latent Concepts}
Figure~\ref{fig:wordclouds_roberta} presents examples of latent concepts learned by the RoBERTa model on the AG News dataset, with two concepts drawn from each of the \textit{Sports}, \textit{Sci/Tech}, and \textit{World}. Within the \textit{Sports} category, Figure~\ref{fig:olympic} and Figure~\ref{fig:baseball_roberta} show two distinct concepts. Figure~\ref{fig:olympic} captures Olympic sports, with terms related to swimming and track events such as ``Phelps'', ``freestyle'', ``200-meter'', and ``Thorpe''. Figure~\ref{fig:baseball_roberta} focuses on baseball game recaps, emphasizing team names, inning-level events, and scoring actions, including words such as ``marlines'', ``Cubes'', ``innings'', and ``homered''. For the \textit{Sci/Tech} category, Figure~\ref{fig:biomedical} highlights biomedical research, with concept words such as ``mice'', ``brain'', ``gene'', and ``cancer''. Figure~\ref{fig:computer} captures computer hardware and networking, including terms such as ``Cisco'', ``Dell'', ``laptop'', and ``networking''. Within the \textit{World} category, Figure~\ref{fig:nuclear} focuses on geopolitics and nuclear diplomacy, with words such as ``Pyongyang'', ``nuclear'', ``Tehran'', and ``weapons''. Figure~\ref{fig:Iraq} captures the Shiite uprising in Iraq through specific conflict-related words such as ``Najaf'', ``al-Sadr'', ``cleric'', and ``militants''.

\end{document}